\documentclass{article} 
\usepackage{iclr2026_conference,times}

\usepackage{amsmath,amsfonts,bm}

\def\eqref#1{equation~\ref{#1}}

\def\1{\bm{1}}

\DeclareMathAlphabet{\mathsfit}{\encodingdefault}{\sfdefault}{m}{sl}
\SetMathAlphabet{\mathsfit}{bold}{\encodingdefault}{\sfdefault}{bx}{n}

\usepackage{hyperref}
\usepackage{url}

\usepackage{nicefrac}       
\usepackage{microtype}      
\usepackage[dvipsnames]{xcolor}
\usepackage[most]{tcolorbox}

\usepackage[capitalize,noabbrev]{cleveref}

\usepackage[shortlabels,inline]{enumitem}  
\usepackage{relsize}
\usepackage{graphicx}       
\usepackage{booktabs}       
\usepackage{wrapfig}        
\usepackage{multirow}       
\usepackage{subcaption}
\usepackage{ctable}         
\usepackage[normalem]{ulem} 
\usepackage{colortbl}
\usepackage{algorithmic,algorithm}

\usepackage{wrapfig}
\usepackage{relsize}
\usepackage{graphicx}       
\usepackage{booktabs}       
\usepackage{wrapfig}        
\usepackage{multirow}       

\usepackage{ragged2e} 
\definecolor{mygrey}{RGB}{128, 128, 128}
\definecolor{mygreen}{RGB}{164, 217, 187}
\definecolor{myblue}{RGB}{105, 158, 212}
\definecolor{myred}{RGB}{239, 129, 131}
\definecolor{myorange}{RGB}{255, 181, 112}
\definecolor{mypurple}{RGB}{195, 171, 208}
\definecolor{codegreen}{RGB}{100, 143, 79}

\newtcbox{\greytb}{on line,
  colback=mygrey, colframe=mygrey,
  boxrule=0.8pt, arc=4pt, boxsep=2pt,
  left=0.5pt, right=0.5pt, top=0.5pt, bottom=0.5pt,
  fontupper=\color{black}, enhanced}
\newtcbox{\greentb}{on line,
    colback=mygreen, colframe=mygreen,
    boxrule=0.8pt, arc=4pt, boxsep=2pt,
    left=0.5pt, right=0.5pt, top=0.5pt, bottom=0.5pt,
    fontupper=\color{black}, enhanced}
\newtcbox{\bluetb}{on line,
    colback=myblue, colframe=myblue,
    boxrule=0.8pt, arc=4pt, boxsep=2pt,
    left=0.5pt, right=0.5pt, top=0.5pt, bottom=0.5pt,
    fontupper=\color{black}, enhanced}
\newtcbox{\redtb}{on line,
    colback=myred, colframe=myred,
    boxrule=0.8pt, arc=4pt, boxsep=2pt,
    left=0.5pt, right=0.5pt, top=0.5pt, bottom=0.5pt,
    fontupper=\color{black}, enhanced}
\newtcbox{\orangetb}{on line,
    colback=myorange, colframe=myorange,
    boxrule=0.8pt, arc=4pt, boxsep=2pt,
    left=0.5pt, right=0.5pt, top=0.5pt, bottom=0.5pt,
    fontupper=\color{black}, enhanced}
\newtcbox{\purpletb}{on line,   
    colback=mypurple, colframe=mypurple,
    boxrule=0.8pt, arc=4pt, boxsep=2pt,
    left=0.5pt, right=0.5pt, top=0.5pt, bottom=0.5pt,
    fontupper=\color{black}, enhanced}

\title{
\raisebox{-0.017\linewidth}{\includegraphics[width=1cm]{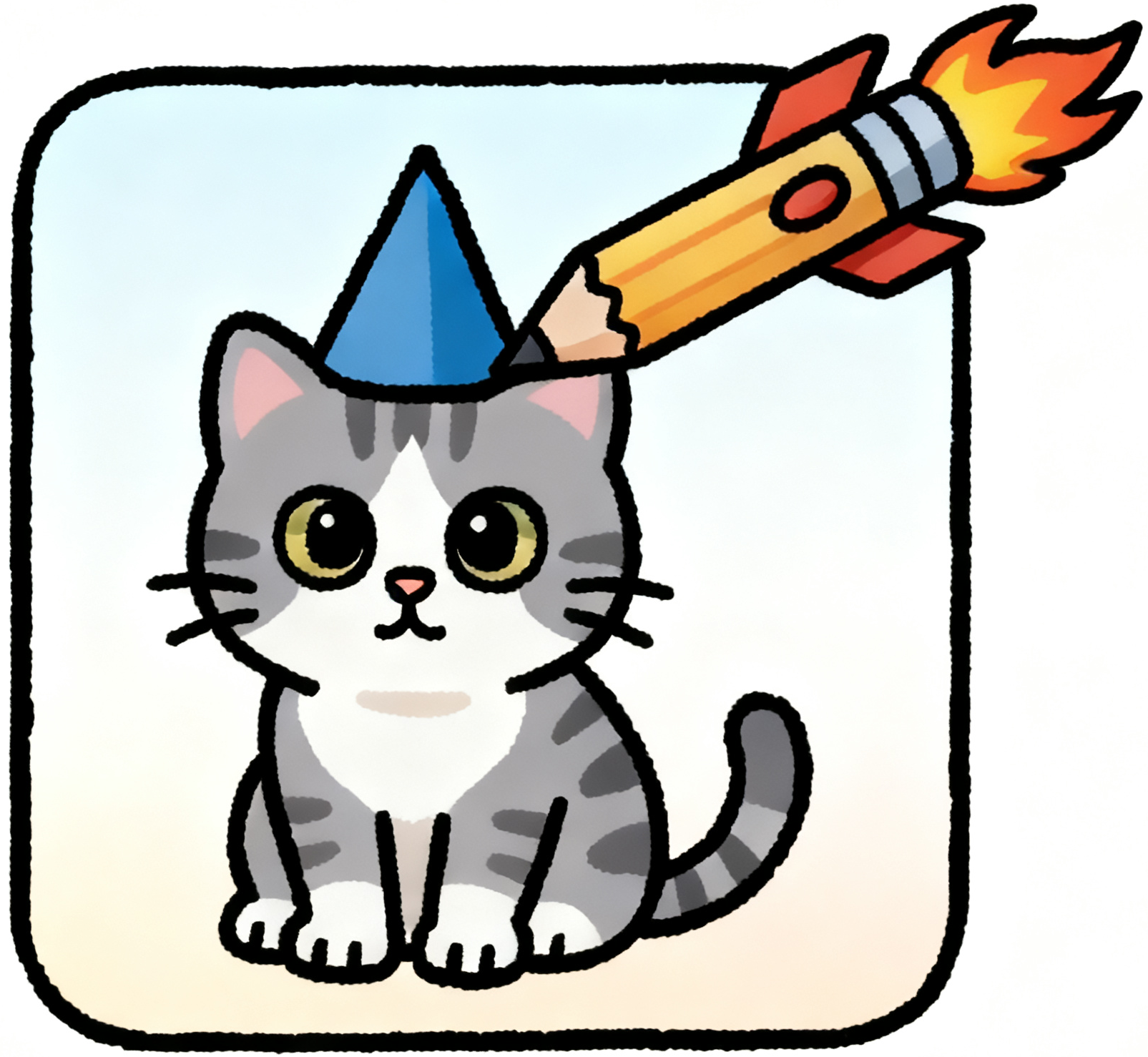}} 
RegionE: Adaptive Region-Aware Generation for Efficient Image Editing}

\author{Pengtao Chen$^{1}$ \quad Xianfang Zeng$^{2 \text{\ddag}}$ \quad Maosen Zhao$^1$ \quad \textbf{Mingzhu Shen}$^3$ \quad Peng Ye$^1$  \\
\textbf{Bangyin Xiang}$^1$ \quad \textbf{Zhibo Wang}$^2$ \quad \textbf{Wei Cheng}$^2$ \quad \textbf{Gang Yu}$^2$ \quad \textbf{Tao Chen}$^{1}$\thanks{\noindent Corresponding author. $^{\ddag}$Project leader. Work was done when interned at StepFun.  \vspace{-15pt}} \\
$^{1}$ Fudan University \quad $^{2}$ StepFun \quad $^{3}$ Imperial College London\\
{\texttt{Code:} \href{https://github.com/Peyton-Chen/RegionE}{\textbf{\texttt{https://github.com/Peyton-Chen/RegionE}}}}
}

\begin{document}

\maketitle

\begin{abstract}

Recently, instruction-based image editing (IIE) has received widespread attention. 
In practice, IIE often modifies only specific regions of an image, while the remaining areas largely remain unchanged. Although these two types of regions differ significantly in generation difficulty and computational redundancy, existing IIE models do not account for this distinction, instead applying a uniform generation process across the entire image. This motivates us to propose \textbf{RegionE}, an adaptive, region-aware generation framework that accelerates IIE tasks without additional training. Specifically, the RegionE framework consists of three main components: \textbf{1) Adaptive Region Partition}. 
We observed that the trajectory of unedited regions is straight, allowing for multi-step denoised predictions to be inferred in a single step. 
Therefore, in the early denoising stages, we partition the image into edited and unedited regions based on the difference between the final estimated result and the reference image. \textbf{2) Region-Aware Generation}.  After distinguishing the regions, we replace multi-step denoising with one-step prediction for unedited areas. 
For edited regions, the trajectory is curved, requiring local iterative denoising. To improve the efficiency and quality of local iterative generation, we propose the Region-Instruction KV Cache, which reduces computational cost while incorporating global information. 
\textbf{3) Adaptive Velocity Decay Cache}. 
Observing that adjacent timesteps in edited regions exhibit strong velocity similarity, we further propose an adaptive velocity decay cache to accelerate the local denoising process.
We applied RegionE to state-of-the-art IIE base models, including Step1X-Edit, FLUX.1 Kontext, and Qwen-Image-Edit. RegionE achieved acceleration factors of 2.57×, 2.41×, and 2.06×, respectively, with minimal quality loss (PSNR: 30.520–32.133). Evaluations by GPT-4o also confirmed that semantic and perceptual fidelity were well preserved.
\end{abstract}

\section{Introduction}
In recent years, diffusion models~\citep{rombach2022diff} have achieved rapid progress in generative tasks, particularly in visual generation, where state-of-the-art models can synthesize highly realistic images. Within this context, the task of editing existing images according to user requirements has gradually emerged as an important direction~\citep{kawar2023imagic}. Recently, diffusion-based foundation models, such as FLUX.1 Kontext~\citep{fluxkontext2025labs}, Qwen-Image-Edit~\citep{qwenimage2025wu}, and Step1X-Edit~\citep{step1x2025liu}, have been developed. These models can perform precise image editing using only textual instructions, offering a novel solution for instruction-based image editing and providing more powerful tools for image post-processing~\citep{choi2024tryon}.

Although diffusion-based IIE models can achieve impressive editing results, their high inference latency limits their use in real-time applications. Previous research on efficient diffusion inference has primarily focused on image generation. For instance, some studies reduce model parameters through pruning~\citep{rombach2022diff,castells2024ldopruner}, others decrease model bit-width via quantization~\citep{shang2023ptq4dm,zhao2025pioneering}, and some employ distillation to reduce model size~\citep{kim2023bksdm} and the number of timesteps~\citep{sauer2024add}. In the two-stage inversion-based editing paradigm~\citep{pan2023effectiverealimageediting,wang2025tamingrectifiedflowinversion}, redundancy in the inversion and denoising stages has been analyzed, leading to methods like EEdit~\citep{yan2025eedit} that accelerate both stages simultaneously. However, for the emerging denoising-only paradigm of IIE, the redundancy and feasibility of efficient inference remain largely unexplored.

\begin{figure*}[t]
    \centering

    \includegraphics[width=0.90 \linewidth]{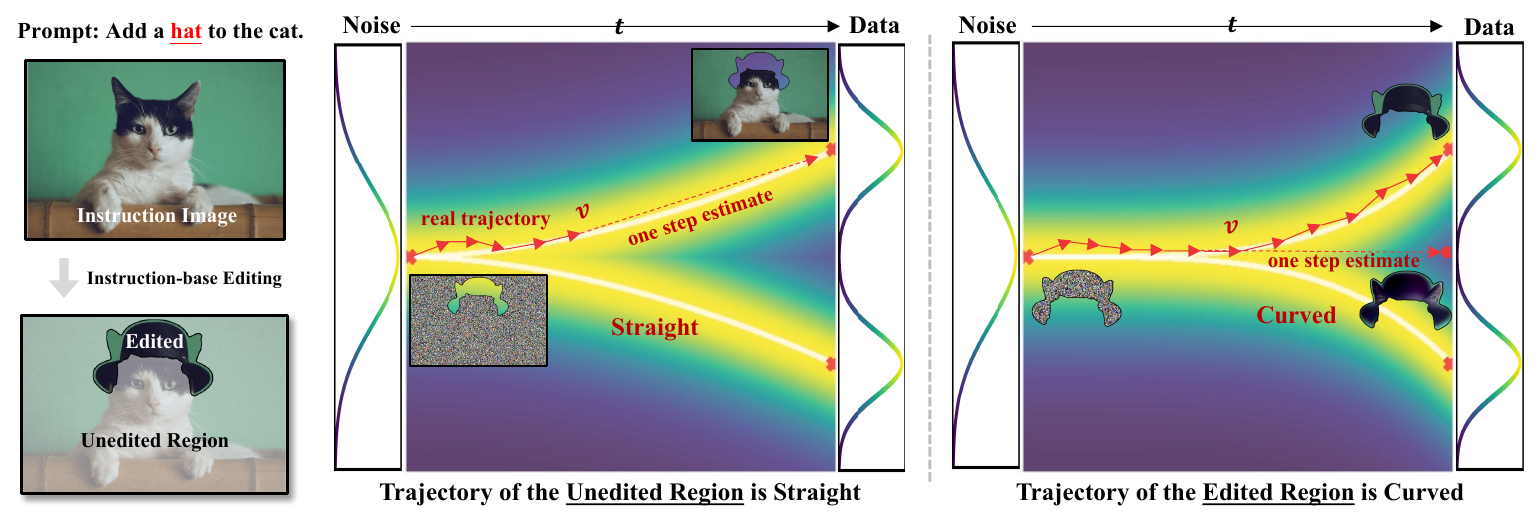}
    \caption{Trajectories of different regions in the IIE task. In unedited regions, the trajectory is nearly linear, allowing early-stage velocity to provide a reliable estimate of the multi-step denoised images, including the final result. In contrast, edited regions exhibit curved trajectories, making the final image harder to predict. Despite this, the velocity between consecutive timesteps remains consistent.}
    \vspace{-11pt} 
    \label{fig:teasor}
\end{figure*}

Our study reveals that current IIE models exhibit two significant types of redundancy: 1) Spatial Generation Redundancy. Unlike image generation tasks, which require reconstructing the entire image, IIE models often need to modify only local regions specified by the instructions, while the remaining areas remain essentially unchanged. For example, as shown in Figure~\ref{fig:teasor}, the model edits only the region around the hat. Nevertheless, IIE models apply the same computational effort to both edited and unedited areas, resulting in significant redundancy in the latter. 
2) Redundancy across diffusion timesteps. First, at neighboring timesteps, the key and value within the attention layers at the same network depth are highly similar. Second, in the middle stages of denoising, the velocity output by the diffusion transformer (DiT) at adjacent timesteps is also highly similar.

To mitigate spatial and temporal redundancy in IIE models, this paper introduces RegionE, a training-free, adaptive, and region-aware generative framework that accelerates the current IIE models. 
Firstly, we observed that the trajectories of edited regions are often more curved, making it difficult to accurately predict the final edited results at early timesteps, as shown in Figure~\ref{fig:teasor}. In contrast, unedited regions follow nearly linear trajectories, allowing more reliable predictions from the same early steps. 
Based on this observation, RegionE introduces an Adaptive Region Partition (ARP), which performs a one-step estimation for the final image in the early stage and compares its similarity with the reference (instruction) image. Regions with high similarity (minimal change after editing) are classified as unedited, whereas regions with low similarity are classified as edited.
Then, we perform region-aware generation on the two separated parts. Specifically,
We replace multi-step denoising with one-step estimation for the unedited areas and apply region-iterative denoising for edited areas. 
During edited region generation, RegionE discards unedited region tokens and instruction image tokens, and effectively reinjects global context into local generation through our proposed Region-Instruction KV Cache (RIKVCache), which leverages the similarity of key and value across timesteps. This process primarily addresses redundancy in spatial. 
Finally, regarding temporal redundancy, we find that the velocity outputs of DiT at adjacent timesteps are highly consistent in direction but decay in magnitude over time, with the decay dependent on the timestep. 
To exploit this property, RegionE introduces an Adaptive Velocity Decay Cache (AVDCache), which accurately models this pattern and further accelerates the region generation process. 
Experimental results demonstrate that RegionE achieves speedups of approximately 2.57×, 2.41×, and 2.06× on Step1X-Edit, FLUX.1 Kontext, and Qwen-Image-Edit, respectively, while maintaining PSNR values of 30.520, 32.133, and 31.115 before and after acceleration. Evaluations using GPT-4o further indicate that the perceptual differences are negligible, confirming that RegionE effectively eliminates redundancy in IIE tasks without compromising image quality.

The contributions of our paper are as follows:

\begin{itemize}[leftmargin=1em]
\item We observe that in IIE tasks, unedited regions exhibit nearly linear generation trajectories, allowing early-stage velocities to provide reliable estimates for multi-step denoised images, including the final image. In contrast, edited regions follow more curved trajectories, making the final image harder to predict. Nevertheless, the velocity remains consistent across consecutive timesteps.

\item We propose RegionE, a training-free, efficient IIE method with adaptive, region-aware generation. It reduces spatial redundancy by performing early adaptive predictions for edited and unedited regions and generating each region locally in subsequent stages, while mitigating temporal redundancy via a velocity-decay cache across timesteps.

\item RegionE achieves 2.57×, 2.41×, and 2.06× end-to-end speedups on Step1X-Edit, FLUX.1 Kontext, and Qwen-Image-Edit, while maintaining PSNR (30.520, 32.133, 31.115) and SSIM (0.939, 0.917, 0.937). Evaluations with GPT-4o further confirm that no quality degradation occurs.
\end{itemize}


\section{Related Work}

\textbf{Efficient Diffusion Model.}
Although few efficient methods have been developed specifically for IIE models, a variety of acceleration techniques have been proposed for diffusion models more generally. From the perspective of parameter redundancy, researchers have introduced pruning methods such as Diff-Pruning~\cite{fang2023diffpruning} and LD-Pruner~\citep{castells2024ldopruner}, quantization methods such as PTQ4DM~\citep{shang2023ptq4dm}, FPQuant~\citep{zhao2025pioneering}, and SVDQuant~\citep{li2024svdquant}, distillation methods such as BK-SDM~\citep{kim2023bksdm} and CLEAR~\citep{liu2024clear}, sparse attention methods such as DiTFastAttn~\citep{yuan2025ditfastattn, zhang2025ditfastattnv2}, SVG~\citep{xi2025svg}, Sparse-vDiT~\citep{chen2025sparsevdit}, and VORTA~\citep{sun2025vorta}, and early-stopping strategies such as ES-DDPM~\citep{lyu2022earlystop}. From the perspective of temporal redundancy, methods like DeepCache~\citep{ma2024deepcache}, $\Delta$-DiT~\citep{chen2024deltadit}, FORA~\citep{selvaraju2024fora}, and TeaCache~\citep{liu2025teacache} reuse intermediate features across timesteps~\citep{shen2024mddit, liu2025speca, liu2025taylor}, while approaches such as LCM~\citep{luo2023lcm} and ADD~\citep{sauer2024add} reduce the number of timesteps through model distillation. 
From the perspective of spatial redundancy, RAS~\citep{liu2025ras} observes that at each diffusion timestep, the model may focus only on semantically coherent regions; therefore, only those regions need to be updated, thereby accelerating image generation. Similarly, ToCa~\citep{zou2024toca} and DuCa~\citep{zou2024duca} note that during denoising, different tokens exhibit varying sensitivities, and dynamically updating only a subset of tokens at each timestep can further accelerate image generation.
In contrast to the methods above, RegionE leverages the trajectory characteristics unique to IIE tasks, while simultaneously addressing both spatial and temporal redundancies in diffusion-based image editing to achieve acceleration.

\textbf{Image Editing.}
Image editing is an essential task in the field of generative modeling. In the early U-Net~\citep{unet2015ronneberger} era, ControlNet~\citep{controlnet2023zhang} introduced a robust editing solution through a repeat-structure design. As research advanced, inversion-based methods~\citep{pan2023effectiverealimageediting,wang2025tamingrectifiedflowinversion} gradually became the dominant approach. These methods apply noise to the original image in the latent space and then recover the edited result through a denoising process. However, this paradigm involves both inversion and denoising stages, which increases complexity. At the same time, IIE models began to emerge. Approaches such as InstructEdit~\citep{wang2023instructedit}, MagicBrush~\citep{zhang2023magicbrush}, and BrushEdit~\citep{li2024brushedit} employed modular pipelines, in which large language models generate prompts, spatial cues, or synthetic instruction–image pairs to guide diffusion-based editing. Most of these approaches, however, are task-specific and lack generality. More recently, a new class of IIE has been developed to improve general-purpose editing. These models rely solely on textual instructions, without requiring masks or task-specific designs, and still achieve effective editing performance. Concretely, they leverage MLLMs or advanced text encoders to provide richer semantic control signals, and feed both the target image and noise into a DiT~\citep{peebles2023dit} architecture to enhance image alignment. In this work, we propose an adaptive, region-aware acceleration method for this emerging IIE models.
\section{Preliminary}

\textbf{Flow Matching \& Rectified Flow.}
Flow matching~\citep{lipman2022flowmatching} has become a widely adopted training technique in advanced diffusion models. It facilitates the transfer from a source distribution $\pi_1$ to a target distribution $\pi_0$ by learning a time-dependent velocity field $\bm v(\bm x,t)$. This velocity field is used to construct the flow through the ordinary differential equation (ODE):
\begin{equation}
    \frac{d\phi_t(\bm x)}{dt}=\bm v(\phi_t(\bm x), t),\phi_1(\bm x)\sim\pi_1.
\end{equation}
Rectified Flow~\citep{liu2022rectifiedflow} simplifies this process through a linear assumption. Given that $\bm X_1$ follows a noise distribution $\pi_1$ and $\bm X_0$ follows the target image distribution $\pi_0$, the equation is 
\begin{equation}
    \bm X_t = (1-t)\bm X_0+t \bm X_1,t\in[0,1].
\end{equation}
Differentiating both ends with respect to timestep $t$ yields: $\frac{d\bm X_t}{dt}=\bm X_1-\bm X_0$. The velocity of the rectified flow $\bm v(\bm X_t, t)$, always points in the direction of $\bm X_1-\bm X_0$. Therefore, the training loss is minimized by reducing the deviation between the velocity and $\bm X_1-\bm X_0$:
\begin{equation}
    \mathcal{L} = \mathbb{E}_{t}\big[||(\bm X_1-\bm X_0)-\bm v(\bm X_t,t)||^2\big].
\end{equation}
The inference process involves starting from $\bm X_1$ and iteratively solving for $\bm X_0$ in reverse, using the learned velocity $\bm v(\bm X_t, t)$. In practice, we typically use a discrete Euler sampler, which discretizes the timestep $t_i (i\in \mathbb{N}^T, t_T=1, t_0=0)$ and approximates:
\begin{equation}
    \bm X_{t_{i-1}} = \bm X_{t_i}-\Delta t_{i,{i-1}}\cdot \bm v(\bm X_{t_i},t_i),\Delta t_{i,{i-1}}=t_{i}-t_{i-1}.\label{eq:inference}
\end{equation}
After $T$ iterations, the final target image $\bm X_0$ is obtained. This paper, therefore, targets the IIE task and optimizes the inference process of $T$ iterations in Equation \ref{eq:inference}.

\begin{figure*}[t]
    \centering
    \includegraphics[width=0.99 \linewidth]{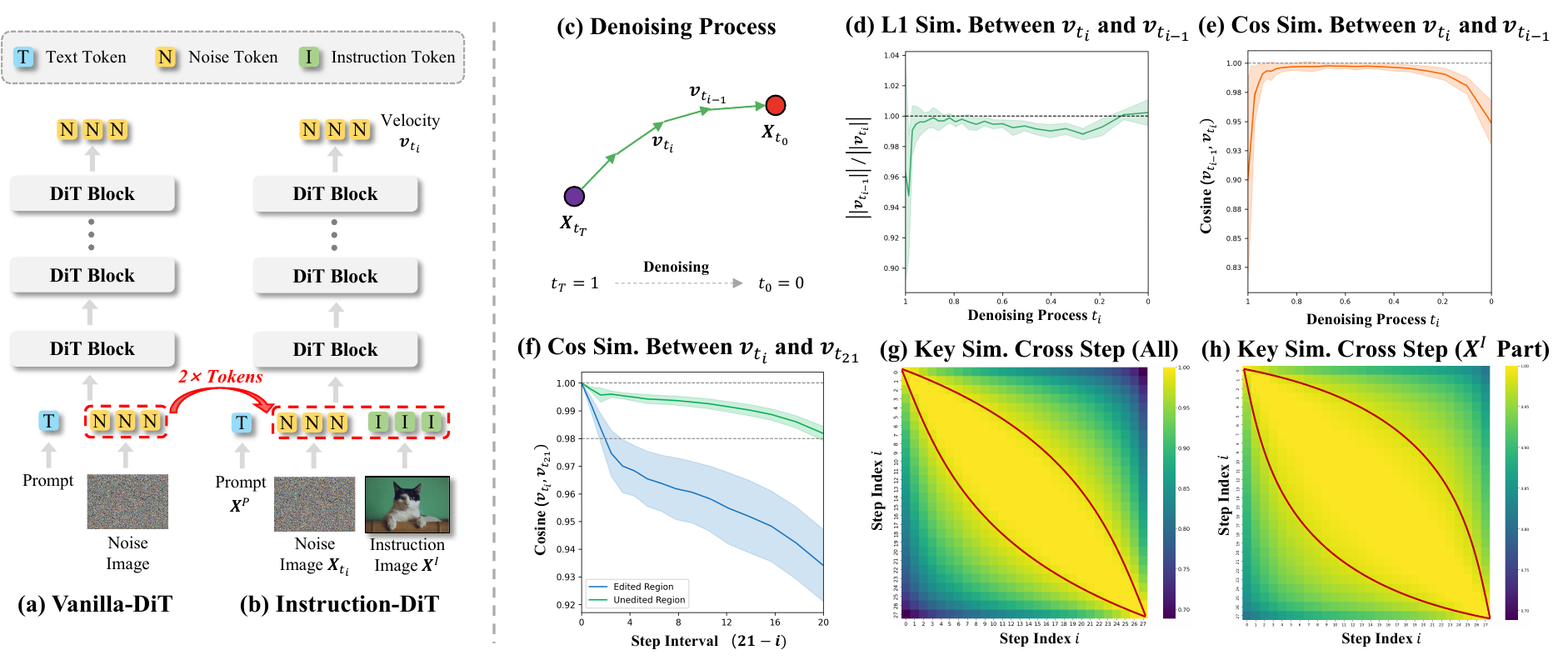}
    \caption{Comparison between traditional DiT and DiT in IIE (a, b). Symbolic visualization of the denoising process (c). L1 and cosine similarities of velocities between adjacent timesteps during denoising (d, e). Cosine similarity between velocities after $t_{21}$ in edited and unedited regions with $\bm v_{21}$ (f). Cross-step key similarity (g) and cross-step similarity of instruction-related keys (h).}
    \label{fig:proof}
\end{figure*}
\textbf{Instruction-Based Editing Model.}
Recent IIE models, such as Step1X-Edit~\citep{step1x2025liu}, FLUX.1 Kontext~\citep{fluxkontext2025labs} and Qwen-Image~\citep{qwenimage2025wu}, follow the same paradigm, as shown in Figure~\ref{fig:proof}b. In these models, the velocity field is estimated using Instruction-DiT, the variants of DiT~\citep{peebles2023dit}. The input to Instruction-DiT consists of three types of tokens: text (prompt) tokens $\bm X^{P}$, noise tokens $\bm X_{t_i}$, and instruction tokens $\bm X^I$. The noise token corresponds to the generation of the target image, while the text token carries the instruction information. The instruction token is specific to the editing task, representing the part of the image to be edited. Notably, the counts of noise and instruction tokens are roughly comparable and substantially higher than that of text tokens. Temporally, the text and instruction tokens serve as static control signals throughout the denoising process, whereas the noise token evolves dynamically at each timestep. Since Instruction-DiT is designed to predict only the noise component, the model’s output corresponds exclusively to the portion represented by the noise token. To simplify the expression, the Instruction-DiT mentioned below will be referred to simply as DiT.

\section{Methodology}
\label{sec:method}

\begin{figure*}[t]
    \centering
    \includegraphics[width=0.99 \linewidth]{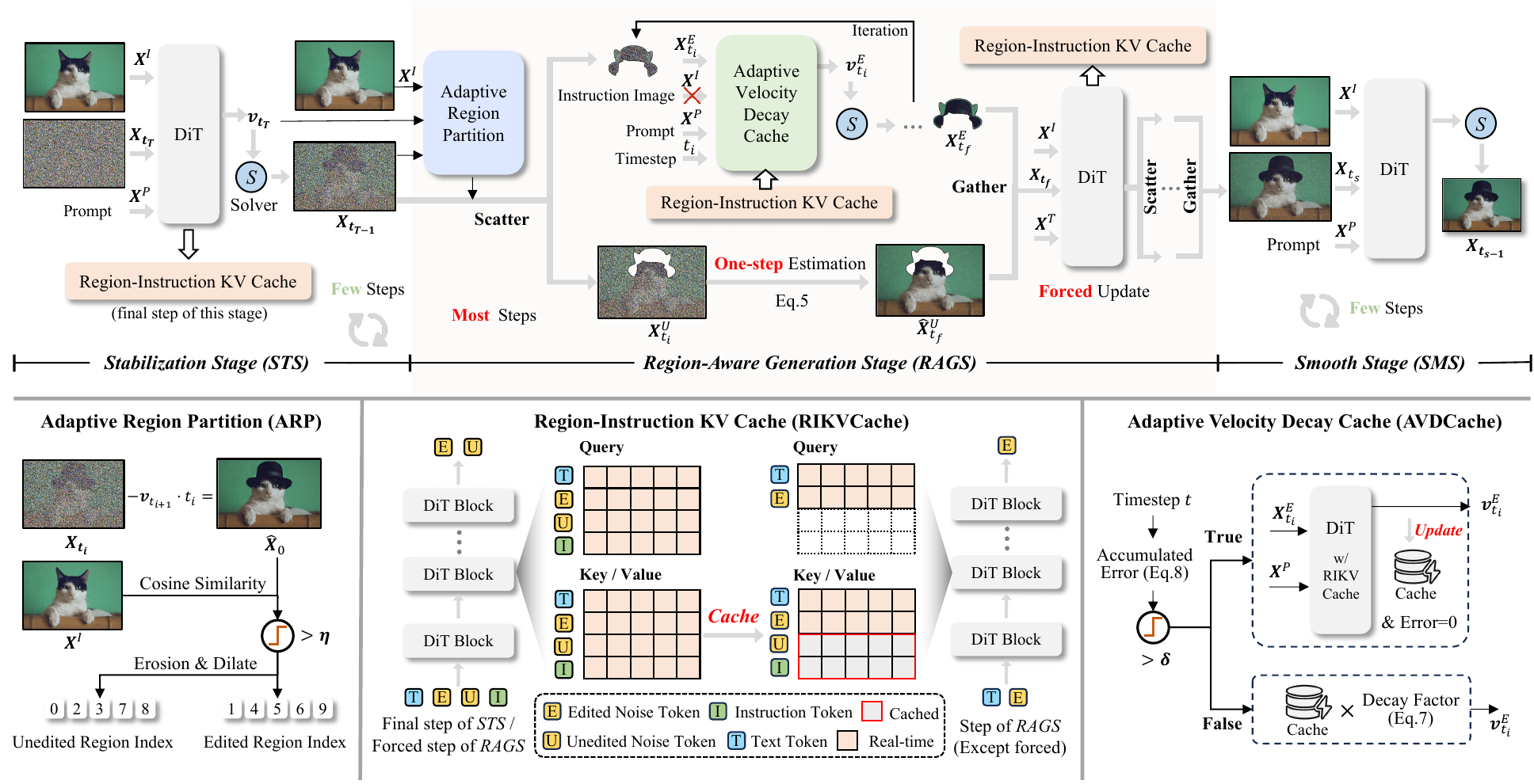}
    \caption{\textbf{Overview of the RegionE}. RegionE consists of three stages: STS, RAGS, and SMS. In the STS, no acceleration is applied due to unstable DiT outputs, and all KV values are cached at the final step. In the RAGS, an Adaptive Region Partition distinguishes between edited and unedited regions: unedited regions are denoised in one step, while edited regions are generated iteratively. This iterative generation process leverages RIKVCache for injecting global information and AVDCache for acceleration. Certain forced-update steps aggregate the full image to refresh RIKVCache with complete DiT computation. Finally, in the SMS, several full denoising steps are performed to eliminate artifacts along the boundaries between edited and unedited regions.}
    \label{fig:pipeline}
\end{figure*}
This section introduces RegionE, a method that accelerates the IIE model without additional training. The workflow is shown in Figure \ref{fig:pipeline}. RegionE consists of three stages: the Stabilization Stage (STS), the Region-Aware Generation Stage (RAGS), and the Smooth Stage (SMS).

\textbf{Stabilization Stage.}
In the early steps of denoising, the input $\bm X_{t_{i}}$ to DiT is close to Gaussian noise (i.e., the signal-to-noise ratio is low). This leads to oscillations in DiT’s velocity estimation (see Figure~\ref{fig:proof}d and ~\ref{fig:proof}e). Since the estimates at this stage are inherently unstable, it is not suitable for acceleration. Therefore, we keep the original sampling process unchanged. Additionally, at the last step of this stage, we save the Key and Value in each attention layer of DiT, denoted as $\bm K^C$ and $\bm V^C$.

\textbf{Region-Aware Generation Stage.}
This stage is the core component of RegionE and consists of three parts: adaptive region partition, region-aware generation, and adaptive velocity decay cache. The first two parts primarily address spatial redundancy in IIE, while the third further reduces temporal redundancy across timesteps.

\textbf{Adaptive Region Partition.}
After the stabilization stage, the output of DiT becomes stable. As previously observed, the generation trajectories in the edited regions are curved, whereas those in the unedited regions are straight, as shown in Figure~\ref{fig:teasor} and~\ref{fig:proof}f. Therefore, for the unedited regions $\bm X^U_{t_i}$, we can accurately estimate $\hat{\bm X}^U_{t_f}$ at any timestep 
$t_f (f<i)$ using one-step estimation:
\begin{equation}
    \hat{\bm X}^U_{t_{f}} = \bm X^U_{t_{i}} - \bm v^U(\bm X^U_{t_i},t_i) \cdot \Delta t_{i, f}. 
    \label{eq:onestep}
\end{equation}
When $t_{f}=0$, this corresponds to estimating the final unedited regions $\hat{\bm X}^U_0$, which is nearly identical to the true $\bm X_0^U$. 
However, using Equation~\ref{eq:onestep} for the edited region does not accurately estimate $\hat{\bm X}^E_0$. Based on this difference between the edited and unedited regions, we propose an adaptive region partition (ARP), as illustrated in the lower-left corner of Figure~\ref{fig:pipeline}. Given the velocity $\bm v_{t_{i+1}}$ at the beginning of the region-aware generation stage and the noisy image $\bm X_{t_{i}}$, the final edited result $\hat{\bm X}_0$ can be estimated in one step using Equation~\ref{eq:onestep}. This estimate is reliable in unedited regions but less accurate in edited ones. 
Since the unedited region undergoes minimal change before and after editing, we can compute the cosine similarity between the estimated image $\hat{\bm X_0}$ and the instruction image $\bm X^I$ along the token dimension. Regions with sufficiently high similarity ($>$ threshold $\eta$), that is, small changes before and after editing, are considered unedited regions, while the remainder is treated as the edited region. To account for potential segmentation noise, morphological opening and closing operations are applied to make the two regions more continuous and accurate.

\textbf{Region-Aware Generation.}
After identifying the edited and unedited regions, we apply Equation~\ref{eq:onestep} to the unedited region to directly estimate the denoised image $X^U_{t_f}$ at the next timestep $t_f$ in one step, thereby saving computation for the unedited region. For the edited region, our implementation is as follows: first, the input to DiT is changed from $[\bm X^P, \bm X_{t_i}, \bm X^I]$ to $[\bm X^P, \bm X^E_{t_i}]$, so that DiT only estimates the velocity of the edited region $\bm v^E_{t_i}$. However, since DiT contains attention layers that involve global token interactions, completely discarding the $\bm X^I$ and $\bm X^U_{t_i}$ inputs can gradually inject bias into the estimation of $\bm v^E_{t_i}$ during global attention. To compensate for this loss of information, we propose a Region-Instruction KV Cache (RIKVCache). Specifically, the input to DiT remains $[\bm X^P, \bm X^E_{t_i}]$, but within the attention layers of DiT, it is modified as follows:
\begin{equation}
    softmax(\frac{[\bm Q_P,\bm Q_E]\cdot [\bm K_P,\bm K_E,\bm K^C_U,\bm K^C_I]^T}{\sqrt{d}})\cdot [\bm V_P,\bm V_E,\bm V^C_U,\bm V^C_I].
\end{equation}
The lower corner labels $P$, $E$, $U$, and $I$ represent prompt token, edited region token, unedited region token, and instruction token, respectively. The superscript $C$ in the upper-right corner indicates that the value is taken from the cache of the previous complete computation. And the middle-lower part of Figure~\ref{fig:pipeline} visualizes this process. The feasibility of this approach is supported by the high similarity of the KV pairs between consecutive steps, as shown in Figure~\ref{fig:proof}g and \ref{fig:proof}h.

\textbf{Adaptive Velocity Decay Cache.}
As illustrated in the right part of Figure~\ref{fig:teasor}, although the trajectory of the edited region is curved, the velocities between consecutive timesteps are actually similar. Focusing on the intermediate denoising phase, we observe from Figure~\ref{fig:proof}e that the velocity directions between adjacent steps are almost identical (cosine similarity approaches 1). At the same time, the magnitudes exhibit a gradual decay that varies across timesteps (Figure~\ref{fig:proof}d). Based on this observation, we propose an adaptive velocity decay cache (AVDCache). Specifically, the AVDCache introduces a decay factor:
\begin{equation}
    ||\bm v_{t_{i}}|| / ||\bm v_{t_{i+1}}|| = (1-\Delta t_{t_{i+1},t_{i}})\cdot \gamma_{t_{i}}. 
    \label{eq:decay}
\end{equation}
Here, $(1-\Delta t_{t_{i+1},t_{i}})$ represents the sample-aware component under discrete Euler solver, while $\gamma_{t_{i}}$ represents the timestep-aware component. The solver entirely determines the former, while the latter is obtained by fitting on a randomly sampled dataset. Since the decay factor in Eq.~\ref{eq:decay} characterizes the intrinsic differences between diffusion model timesteps, we introduce the AVDCache criterion:
\begin{equation}
    Criterion = 1-\prod_{i=s}^{e}(1-\Delta t_{t_{i+1},t_{i}})\cdot \gamma_{t_{i}}. \label{eq:accumulate}
\end{equation}
Here,$t_s$ and $t_e$ denote the start and end timesteps of the cache, respectively, while the criterion measures the cumulative error of this process. The decision of whether to apply the cache is made using a threshold $\delta$. The complete process is as follows:
\begin{equation}
    \bm v_{t_{i}}^E = 
    \begin{cases} 
    DiT(\bm X^E_{t_i},\bm X^P) & Criterion > \delta \\
    \bm v^{E,C}_{t_s} \cdot \prod_{m=s}^{i}(1-\Delta t_{t_{m+1},t_{m}})\cdot \gamma_{t_{m}} &  else. \\
    \end{cases}
\end{equation}
The right-lower part of Figure~\ref{fig:pipeline} visualizes this process.
In fact, AVDCache is an improved version of the existing residual cache methods, with further details and analysis provided in the supplementary.

After the above process, we obtain the generated results for both the edited and unedited regions. We then re-gather these results according to their spatial positions to reconstruct the complete image tokens. 
It is worth noting that the similarity of the KV Cache decreases as the timestep increases. To address this issue, we periodically enforce full-image gathering at certain timesteps within the region-aware generation stage, performing a complete DiT computation to update the RIKVCache.

\textbf{Smooth Stage.}
Small gaps may appear at the boundaries between edited and unedited regions after stitching. Although these gaps are often imperceptible in most cases, to ensure the generality of our method, we perform several steps of unaccelerated denoising on the merged full image to smooth these discontinuities. Empirically, two denoising steps are sufficient to eliminate the gaps effectively.

\section{Experiment}
\subsection{Experimental Settings}

\textbf{Pretrained Model \& Dataset.}
We evaluate RegionE on three open-source state-of-the-art IIE models: Step1X-Edit-v1p1~\citep{step1x2025liu}, FLUX.1 Kontext~\citep{fluxkontext2025labs}, and Qwen-Image-Edit~\citep{qwenimage2025wu}. Step1X-Edit adopts a CFG (classifier-free guidance)~\citep{ho2022cfg} scale of 6, FLUX.1 Kontext uses a scale of 2.5, and Qwen-Image-Edit applies a scale of 4. All models are evaluated with 28 sampling steps. For evaluation, we follow the dataset protocols described in the respective technical reports. Specifically, we use 606 image prompt pairs covering 11 tasks from GEdit-Bench English~\citep{step1x2025liu} for Step1X-Edit and Qwen-Image-Edit, and 1026 image prompt pairs spanning five tasks from KontextBench~\citep{fluxkontext2025labs} for FLUX.1 Kontext.

\textbf{Evaluation Metrics.}
We design a comprehensive evaluation framework to assess both the quality and efficiency of IIE models. For quality assessment, we adopt two complementary approaches. First, we evaluate reconstruction quality by measuring deviations before and after acceleration, using PSNR~\citep{zhao2024psnr}, SSIM~\citep{wang2002ssim}, and LPIPS~\citep{zhang2018LPIPS} as metrics. Second, we conduct an editing evaluation using vision–language models (VLMs), specifically GPT-4o, to assess image quality, semantic alignment, and overall performance~\citep{ku2024viescore}, as shown in Table \ref{tab:baseline}. Evaluation dimensions are denoted by the suffixes SC, PQ, and O, consistent with~\citep{step1x2025liu} and~\citep{qwenimage2025wu}. For efficiency evaluation, we report actual runtime latency as well as the relative speedup compared to the vanilla pretrained models.

\textbf{Baseline.}
Currently, there are no acceleration methods designed explicitly for IIEmodels. Therefore, we adapt several effective acceleration techniques initially developed for diffusion models as baselines, since they are also applicable to diffusion-based IIE tasks. From the perspective of timestep redundancy, Steoskip performs larger jumps in the sampling steps, FORA~\citep{selvaraju2024fora} employs block-level cache, and $\Delta$-DiT~\citep{chen2024deltadit} and TeaCache~\citep{liu2025teacache} use residual cache. From the perspective of spatial redundancy, RAS~\citep{liu2025ras} and ToCa~\citep{zou2024toca} perform redundancy-reduction denoising at the token level.

\textbf{Implementation Details.}
For all three models, RegionE uses six steps in the stabilization stage, enforces an update at step 16 in the region-aware generation stage, and adopts two steps in the smooth stage. For Step1X-Edit, FLUX.1 Kontext, and Qwen-Image-Edit, the segmentation thresholds $\eta$ of ARP are 0.88, 0.93, and 0.80, respectively, while the decision thresholds $\delta$ of AVDCache are 0.02, 0.04, and 0.03, respectively. Latency is measured on a single NVIDIA H800 GPU, with each run editing one image at a time. 

\begin{table}
    \caption{Comparison of editing quality and efficiency between RegionE and the baseline. All the evaluations are carried out on a single NVIDIA H800 GPU. \bluetb{S} denotes the strategy for reducing spatial redundancy, while \greentb{T} denotes the strategy for reducing temporal redundancy.}
    \resizebox{\linewidth}{!}{%
    \begin{tabular}{lc|ccc|ccc|cc}
        \toprule[1.5pt]
         \multirow{2}{*}{\textbf{Model}}&  \multirow{2}{*}{\textbf{Type}}&  \multicolumn{3}{c|}{\textbf{Against Vanilla}}& \multicolumn{3}{c|}{\textbf{GPT-4o Score}}&  \multicolumn{2}{c}{\textbf{Efficiency}}\\
 & & \textbf{PSNR$\uparrow$ }& \textbf{SSIM $\uparrow$ }& \textbf{LPIPS $\downarrow$ }& \textbf{G-SC $\uparrow$ }&\textbf{G-PQ $\uparrow$ }& \textbf{G-O $\uparrow$ }& \textbf{Latency (s) $\downarrow$ }&\textbf{Speedup $\uparrow$ }\\
        \midrule
         \multicolumn{2}{l|}{\textbf{Step1X-Edit}~\citep{step1x2025liu}}&  -&  -&  -& 7.479 &7.466 &  6.906 &  27.945 &1.000 
\\
         \; \textbf{+ Stepskip}&  \greentb{T}&  26.719 &  0.898 &  0.096 & 7.491 &7.343 &  6.880 &  12.299 &2.272 
\\
         \; \textbf{+ FORA}~\citep{selvaraju2024fora}&  \greentb{T}&  22.126 &  0.835 &  0.178 & 6.078 &\textbf{7.588}&  5.863 &  14.330 &1.950 
\\
         \; \textbf{+ $\Delta$-DiT}~\citep{chen2024deltadit}&  \greentb{T}&  24.659 &  0.874 &  0.122 & 7.432 &7.233 &  6.795 &  12.728 &2.196 
\\
         \; \textbf{+ TeaCache}~\citep{liu2025teacache}&  \greentb{T}&  28.262 &  0.924 &  0.072 & 7.455 &7.361 &  6.866 &  11.212 &2.493 
\\
         \; \textbf{+ RAS}~\citep{liu2025ras}&  \bluetb{S}&  26.819 &  0.892 &  0.100 & 7.339 &7.072 &  6.615 &  15.239 &1.834 
\\
         \; \textbf{+ ToCa}~\citep{zou2024toca}& \bluetb{S}& 24.699 & 0.844 & 0.152 & 7.185 &6.705 & 6.350 & 22.149 &1.262 
\\
        
        \rowcolor[HTML]{EFEFEF}
         \; \textbf{+ Ours (RegionE)}& \greentb{T} \& \bluetb{S}& \textbf{30.520}& \textbf{0.939}& \textbf{0.054}& \textbf{7.552}&7.405 & \textbf{6.948}& \textbf{10.865}&\textbf{2.572}\\
         \midrule
         \multicolumn{2}{l|}{\textbf{FLUX.1 Kontext}~\citep{fluxkontext2025labs}}& -& -& -& 7.197 &\textbf{6.963} & 6.497 & 14.682 &1.000 
\\
         \; \textbf{+ Stepskip}
& \greentb{T}
& 26.199 & 0.838 & 0.123 & 7.126 &6.938 & 6.463 & 8.512 &1.725 
\\
         \; \textbf{+ FORA}~\citep{selvaraju2024fora}
& \greentb{T}
& 24.685 & 0.809 & 0.146 & 7.085 &6.897 & 6.383 & 7.497 &1.958 
\\
         \; \textbf{+ $\Delta$-DiT}~\citep{chen2024deltadit}
& \greentb{T}
& 20.227 & 0.723 & 
0.225 & 7.055 &6.918 & 6.411 & 6.751 &2.175 
\\
         \; \textbf{+ TeaCache}~\citep{liu2025teacache}
& \greentb{T}
& 28.307 & 0.869 & 
0.097 & 7.233 &6.846 & 6.455 & 6.203 &2.367 
\\
         \; \textbf{+ RAS}~\citep{liu2025ras}
& \bluetb{S}
& 26.217 & 0.829 & 0.132 & 7.216 &6.785 & 6.460 & 8.219 &1.786 
\\
         \; \textbf{+ ToCa}~\citep{zou2024toca}& \bluetb{S}
& 23.906 & 0.767 & 0.192 & 6.985 &6.589 & 6.237 & 11.299 &1.299 
\\
        
        \rowcolor[HTML]{EFEFEF}
         \; \textbf{+ Ours (RegionE)}& \greentb{T} \& \bluetb{S}& \textbf{32.133}& \textbf{0.917}& \textbf{0.057}& \textbf{7.278}&6.953& \textbf{6.538}& \textbf{6.096}&\textbf{2.409}\\
         \midrule
         \multicolumn{2}{l|}{\textbf{Qwen-Image-Edit}~\citep{qwenimage2025wu}}& -& -& -& 8.242 &7.948 & 7.700 & 32.125 &1.000 
\\
         \; \textbf{+ Stepskip}
& \greentb{T}
& 28.439 & 0.892 & 0.077 & 8.090 &7.875 & 7.572 & 17.555 &1.830 
\\
         \; \textbf{+ FORA}~\citep{selvaraju2024fora}
& \greentb{T}
& 26.508 & 0.863 & 0.098 & 8.032 &7.760 & 7.501 & 17.815 &1.803 
\\
         \; \textbf{+ $\Delta$-DiT}~\citep{chen2024deltadit}
& \greentb{T}
& 25.020 & 0.821 & 0.116 & 7.964 &7.718 & 7.417 & 17.470 &1.839 
\\
         \; \textbf{+ TeaCache}~\citep{liu2025teacache}
& \greentb{T}
& 28.314 & 0.900 & 0.075 & 8.084 &7.841 & 7.563 & 16.445 &1.954 
\\
         \; \textbf{+ RAS}~\citep{liu2025ras}
& \bluetb{S}
& 27.251 & 0.879 & 0.090 & 8.152 &7.680 & 7.515 & 22.327 &1.439 
\\
         \; \textbf{+ ToCa}~\citep{zou2024toca}& \bluetb{S}
& OOM& OOM& OOM& OOM&OOM& OOM& OOM&OOM\\
        
         \rowcolor[HTML]{EFEFEF}
         \; \textbf{+ Ours (RegionE)}& \greentb{T} \& \bluetb{S}& \textbf{31.115}& \textbf{0.937}& \textbf{0.046}& \textbf{8.242} &\textbf{7.968} & \textbf{7.731} & \textbf{15.604} &\textbf{2.059} 
\\
         \bottomrule[1.5pt]
    \end{tabular}
    }
    \label{tab:baseline}
\vspace{-11pt}
\end{table}

\subsection{Experimental Results Analysis}
We evaluate RegionE against several state-of-the-art acceleration methods on three prominent IIE models: Step1X-Edit, FLUX.1 Kontext, and Qwen-Image-Edit. Our evaluation encompasses quantitative metrics, efficiency measurements, and visualization, demonstrating that RegionE achieves a superior balance between acceleration and quality preservation. The quantitative results are shown in Table~\ref{tab:baseline}. Since both GEdit-Bench and KontextBench involve multiple editing tasks, the table reports results averaged over tasks, while the per-task quantitative results are provided in the supplementary.

\textbf{Deviation Analysis Compared to Pre-trained Models.}
The "Against Vanilla" evaluation reveals RegionE's exceptional fidelity to original model outputs across all evaluation metrics, significantly outperforming competing acceleration methods. RegionE achieves the highest PSNR values: 30.520 dB (Step1X-Edit), 32.133 dB (FLUX.1 Kontext), and 31.115 dB (Qwen-Image-Edit), representing substantial improvements of 2-4 dB over the next-best methods, indicating minimal pixel-level deviation from the original outputs. The SSIM scores of 0.939, 0.917, and 0.937 demonstrate superior preservation of structural coherence across different model architectures. In contrast, the LPIPS scores of 0.054, 0.057, and 0.046 represent 25-50\% improvements over competing methods, indicating dramatically reduced perceptual differences that would be virtually indistinguishable to users. This consistent performance across three diverse model architectures validates RegionE's architectural agnosticism. RegionE consistently maintains stable, high-quality results.

\textbf{GPT-4o Editing Quality Assessment.}
The GPT-4o evaluation provides additional quality validation through automated semantic and perceptual analysis across three dimensions, consistently demonstrating RegionE's superior performance. For semantic consistency (G-SC), RegionE achieves scores of 7.552, 7.278, and 8.242, matching or exceeding original models while maintaining substantial acceleration, with Qwen-Image-Edit showing perfect preservation (8.242) despite 2.059× speedup. The perceptual quality (G-PQ) scores of 7.405, 6.953, and 7.968 consistently outperform competing acceleration methods by 0.1 to 0.3 points, demonstrating the practical preservation of visual coherence through region-aware processing. Overall quality (G-O) scores of 6.948, 6.538, and 7.731 provide holistic assessment validation, with the alignment between GPT-4o assessments and quantitative metrics (PSNR, SSIM, LPIPS) strengthening confidence in RegionE's comprehensive quality preservation across multiple evaluation dimensions and providing additional evidence of the hybrid temporal-spatial optimization approach's effectiveness.

\textbf{Efficiency Analysis.}
RegionE demonstrates substantial efficiency gains while maintaining superior quality, achieving an optimal balance between acceleration and performance preservation with impressive results across all evaluated models. The method achieves speedups of 2.572×, 2.409×, and 2.059× across Step1X-Edit, FLUX.1 Kontext, and Qwen-Image-Edit respectively, translating to significant absolute latency reductions: from 27.945s to 10.865s, from 14.682s to 6.096s, and from 32.125s to 15.604s respectively. RegionE occupies the optimal position on the efficiency-quality curve, maintaining the highest quality metrics while achieving competitive or superior acceleration compared to methods that sacrifice substantial quality for higher speedups.  

\begin{figure*}[t]
    \centering
    \includegraphics[width=0.95 \linewidth]{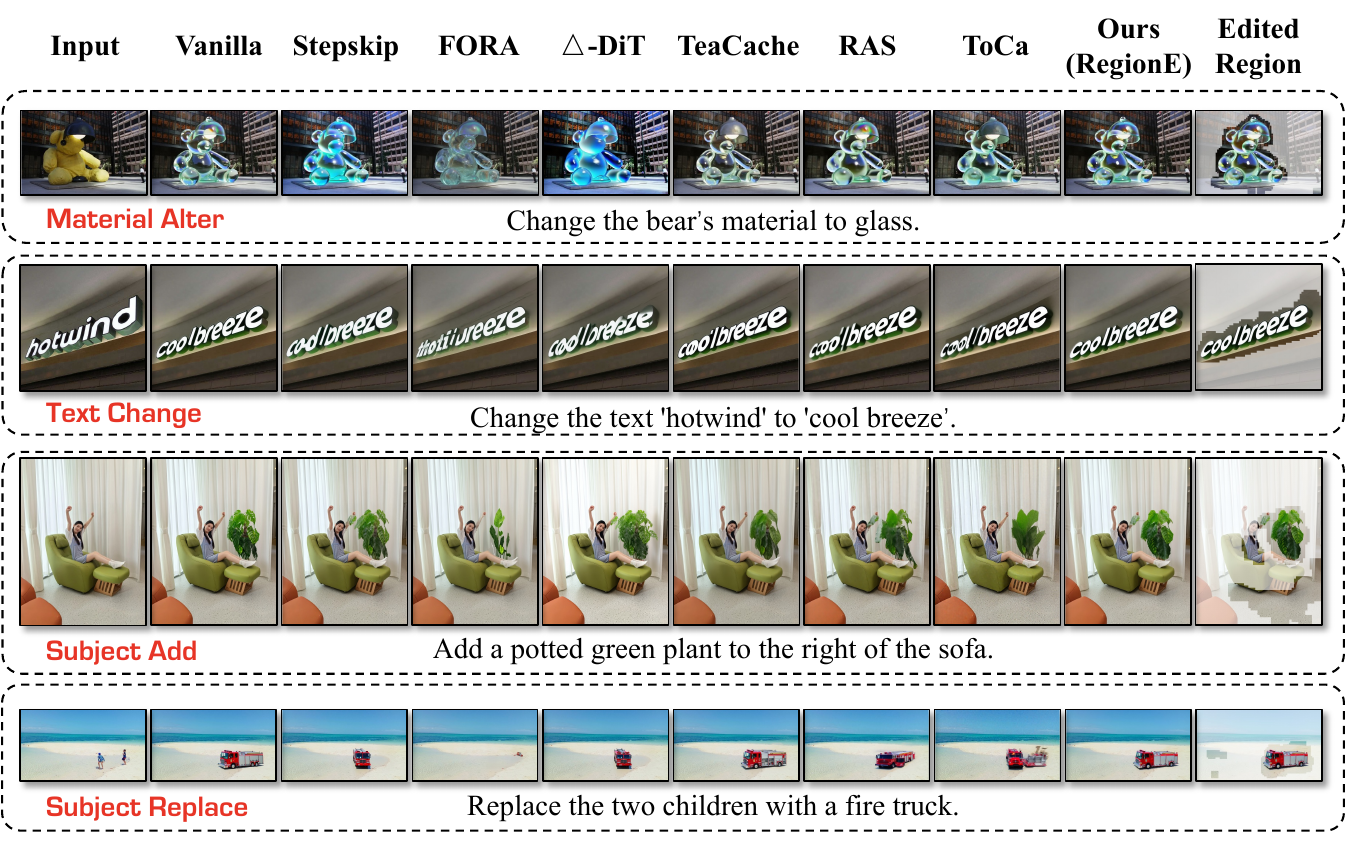}
    \caption{Examples of edited images by RegionE and baseline on Step1X-Edit-v1p1.}
    \label{fig:visu}
    \vspace{-10pt}
\end{figure*}

\textbf{Visualization.}
Figure~\ref{fig:visu} presents partial visualizations of different acceleration methods on  Step1X-Edit. Among the baselines, RegionE produces edited outputs closest to the vanilla setting at higher speedups, preserving both details and contours. The last column shows ARP predictions of spatial regions in RegionE, where unedited regions are masked. These masked regions closely match human perception. Additional visualizations for other tasks and models are provided in the supplementary.

The experimental results conclusively demonstrate the effectiveness of RegionE in addressing both spatial and temporal redundancies in IIE models. RegionE achieves superior quality preservation metrics while maintaining competitive acceleration. The consistent improvements across diverse model architectures validate the generalizability of the underlying insights about regional editing patterns and temporal similarities in diffusion-based IIE processes.

\subsection{Ablation Study}
\begin{table}
    \caption{Ablation study on cache design and stage design in RegionE.}
    \resizebox{\linewidth}{!}{%
    \begin{tabular}{ll|ccc|ccc|cc}
        \toprule[1.5pt]
          \multicolumn{2}{c|}{\multirow{2}{*}{\textbf{Variant}}}&  \multicolumn{3}{c|}{\textbf{Against Vanilla}}& \multicolumn{3}{c|}{\textbf{GPT-4o Score}}&  \multicolumn{2}{c}{\textbf{Efficiency}}\\
  \multicolumn{2}{c}{}& \textbf{PSNR$\uparrow$ }& \textbf{SSIM $\uparrow$ }& \textbf{LPIPS $\downarrow$ }& \textbf{G-SC $\uparrow$ }&\textbf{G-PQ $\uparrow$ }& \textbf{G-O $\uparrow$ }& \textbf{Latency (s) $\downarrow$ }&\textbf{Speedup $\uparrow$ }\\
        \midrule
            \rowcolor[HTML]{EFEFEF}
          \multicolumn{2}{c|}{\textbf{RegionE}}&  30.520&  0.939&  0.054& 7.552&7.405&  6.948&  10.865&2.572\\
\midrule
          
          \multirow{2}{*}{\textbf{Cache Design}}&w/o RIKVCache&  22.868&  0.822&  0.207& 5.997&5.389&  5.191
&  10.223&2.734
\\
          &w/o  AVDCache&  31.139&  0.946&  0.046& 7.570&7.482&  7.023
&  16.122&1.733
\\
\midrule
          \multirow{3}{*}{\textbf{Stage Design}}&w/o STS&  21.441&  0.814&  0.161& 7.045&6.758&  6.325
&  7.149&3.909
\\
          &w/o SMS&  28.857&  0.904&  0.085& 7.456&7.207&  6.773
&  9.766&2.862
\\
          &w/o Forced Step&  28.452&  0.915&  0.080& 7.536&7.305&  6.925
&  10.204&2.739 
\\
    \bottomrule[1.5pt]
    \end{tabular}
    }
    \vspace{-10pt}
    \label{tab:ablation}
\end{table}

We conduct ablation studies to investigate the contributions of different components in RegionE, primarily on the Step1X-Edit-v1p1. The quantitative results are summarized in Table~\ref{tab:ablation}.

\textbf{Cache Design.}
We propose two key components: RIKVCache and AVDCache. Removing RIKVCache, i.e., performing local attention within the edited region without injecting instruction information or context from the unedited region, results in a 2.734× speed-up. However, this comes at a significant cost to editing quality, with PSNR dropping from 30.520 to 22.868 and G-O decreasing from 6.948 to 5.191. This demonstrates that global context supervision is crucial even during region generation. In contrast, removing AVDCache results in a slight improvement in editing quality (G-O increases from 6.948 to 7.023), but without eliminating redundancy across timesteps, the acceleration is limited to 1.733. This indicates that AVDCache significantly improves inference efficiency with minimal degradation in quality.

\textbf{Stage Design.}
We introduce two auxiliary stages: Stabilization Stage (STS) and Smooth Stage (SMS), as well as a forced step in the region-aware generation stage (RAGS). Removing STS causes substantial drops in editing quality (PSNR: 30.520 → 21.441, LPIPS: 0.054 → 0.161, G-O: 6.948 → 6.325). As discussed in Section~\ref{sec:method}, STS addresses the instability in speed estimation, and skipping it results in degraded performance. Removing SMS leads to smaller declines in both pixel-level (PSNR: 30.520 → 28.857, SSIM: 0.939 → 0.904) and perceptual metrics (G-O: 6.948 → 6.773), reflecting its role in bridging the gap between edited and unedited regions. Finally, when the forced step in RAGS was removed, since its role was to mitigate the decay of KV similarity over time, its removal led to a 2-point drop in PSNR, further validating its necessity.

\section{Conclusion}

Inspired by temporal and spatial redundancy in IIE, we propose RegionE, an adaptive, region-aware generation framework that accelerates the IIE process. Specifically, we perform early prediction on spatial regions using ARP and combine it with RIKVCache for region-wise editing to reduce spatial redundancy. We also use AVDCache to minimize temporal redundancy. Experiments show that RegionE achieves 2.57×, 2.41×, and 2.06× end-to-end speedups on Step1X-Edit and FLUX.1 Kontext, and Qwen-Image-Edit, respectively, while maintaining minimal bias (PSNR 30.52–32.13) and negligible quality loss (GPT-4o evaluation results remain comparable). These results demonstrate the effectiveness of RegionE in reducing redundancy in IIE.

\bibliography{iclr2026_conference}

\begin{thebibliography}{49}
\providecommand{\natexlab}[1]{#1}
\providecommand{\url}[1]{\texttt{#1}}
\expandafter\ifx\csname urlstyle\endcsname\relax
  \providecommand{\doi}[1]{doi: #1}\else
  \providecommand{\doi}{doi: \begingroup \urlstyle{rm}\Url}\fi

\bibitem[Bu et~al.(2025)Bu, Ling, Zhou, Wang, Zang, Wu, Lin, and Wang]{bu2025dicacheletdiffusionmodel}
Jiazi Bu, Pengyang Ling, Yujie Zhou, Yibin Wang, Yuhang Zang, Tong Wu, Dahua Lin, and Jiaqi Wang.
\newblock Dicache: Let diffusion model determine its own cache, 2025.

\bibitem[Castells et~al.(2024)Castells, Song, Kim, and Choi]{castells2024ldopruner}
Thibault Castells, Hyoung-Kyu Song, Bo-Kyeong Kim, and Shinkook Choi.
\newblock Ld-pruner: Efficient pruning of latent diffusion models using task-agnostic insights.
\newblock In \emph{Proceedings of the IEEE/CVF Conference on Computer Vision and Pattern Recognition}, pp.\  821--830, 2024.

\bibitem[Chen et~al.(2024)Chen, Shen, Ye, Cao, Tu, Bouganis, Zhao, and Chen]{chen2024deltadit}
Pengtao Chen, Mingzhu Shen, Peng Ye, Jianjian Cao, Chongjun Tu, Christos-Savvas Bouganis, Yiren Zhao, and Tao Chen.
\newblock $\delta$-dit: A training-free acceleration method tailored for diffusion transformers.
\newblock \emph{arXiv preprint arXiv:2406.01125}, 2024.

\bibitem[Chen et~al.(2025)Chen, Zeng, Zhao, Ye, Shen, Cheng, Yu, and Chen]{chen2025sparsevdit}
Pengtao Chen, Xianfang Zeng, Maosen Zhao, Peng Ye, Mingzhu Shen, Wei Cheng, Gang Yu, and Tao Chen.
\newblock Sparse-vdit: Unleashing the power of sparse attention to accelerate video diffusion transformers, 2025.

\bibitem[Choi et~al.(2024)Choi, Kwak, Lee, Choi, and Shin]{choi2024tryon}
Yisol Choi, Sangkyung Kwak, Kyungmin Lee, Hyungwon Choi, and Jinwoo Shin.
\newblock Improving diffusion models for authentic virtual try-on in the wild.
\newblock In \emph{European Conference on Computer Vision}, pp.\  206--235. Springer, 2024.

\bibitem[Fang et~al.(2023)Fang, Ma, and Wang]{fang2023diffpruning}
Gongfan Fang, Xinyin Ma, and Xinchao Wang.
\newblock Structural pruning for diffusion models.
\newblock In \emph{Advances in Neural Information Processing Systems}, 2023.

\bibitem[Ho \& Salimans(2022)Ho and Salimans]{ho2022cfg}
Jonathan Ho and Tim Salimans.
\newblock Classifier-free diffusion guidance.
\newblock \emph{arXiv preprint arXiv:2207.12598}, 2022.

\bibitem[Kawar et~al.(2023)Kawar, Zada, Lang, Tov, Chang, Dekel, Mosseri, and Irani]{kawar2023imagic}
Bahjat Kawar, Shiran Zada, Oran Lang, Omer Tov, Huiwen Chang, Tali Dekel, Inbar Mosseri, and Michal Irani.
\newblock Imagic: Text-based real image editing with diffusion models.
\newblock In \emph{Proceedings of the IEEE/CVF conference on computer vision and pattern recognition}, pp.\  6007--6017, 2023.

\bibitem[Kim et~al.(2023)Kim, Song, Castells, and Choi]{kim2023bksdm}
Bo-Kyeong Kim, Hyoung-Kyu Song, Thibault Castells, and Shinkook Choi.
\newblock Bk-sdm: Architecturally compressed stable diffusion for efficient text-to-image generation.
\newblock In \emph{Workshop on Efficient Systems for Foundation Models@ ICML}, 2023.

\bibitem[Ku et~al.(2024)Ku, Jiang, Wei, Yue, and Chen]{ku2024viescore}
Max Ku, Dongfu Jiang, Cong Wei, Xiang Yue, and Wenhu Chen.
\newblock Viescore: Towards explainable metrics for conditional image synthesis evaluation, 2024.

\bibitem[Labs et~al.(2025)Labs, Batifol, Blattmann, Boesel, Consul, Diagne, Dockhorn, English, English, Esser, et~al.]{fluxkontext2025labs}
Black~Forest Labs, Stephen Batifol, Andreas Blattmann, Frederic Boesel, Saksham Consul, Cyril Diagne, Tim Dockhorn, Jack English, Zion English, Patrick Esser, et~al.
\newblock Flux. 1 kontext: Flow matching for in-context image generation and editing in latent space.
\newblock \emph{arXiv preprint arXiv:2506.15742}, 2025.

\bibitem[Li et~al.(2024{\natexlab{a}})Li, Lin, Zhang, Cai, Li, Guo, Xie, Meng, Zhu, and Han]{li2024svdquant}
Muyang Li, Yujun Lin, Zhekai Zhang, Tianle Cai, Xiuyu Li, Junxian Guo, Enze Xie, Chenlin Meng, Jun-Yan Zhu, and Song Han.
\newblock Svdquant: Absorbing outliers by low-rank components for 4-bit diffusion models.
\newblock \emph{arXiv preprint arXiv:2411.05007}, 2024{\natexlab{a}}.

\bibitem[Li et~al.(2024{\natexlab{b}})Li, Bian, Ju, Zhang, Zhuang, Shan, Zou, and Xu]{li2024brushedit}
Yaowei Li, Yuxuan Bian, Xuan Ju, Zhaoyang Zhang, Junhao Zhuang, Ying Shan, Yuexian Zou, and Qiang Xu.
\newblock Brushedit: All-in-one image inpainting and editing.
\newblock \emph{arXiv preprint arXiv:2412.10316}, 2024{\natexlab{b}}.

\bibitem[Lipman et~al.(2022)Lipman, Chen, Ben-Hamu, Nickel, and Le]{lipman2022flowmatching}
Yaron Lipman, Ricky~TQ Chen, Heli Ben-Hamu, Maximilian Nickel, and Matt Le.
\newblock Flow matching for generative modeling.
\newblock \emph{arXiv preprint arXiv:2210.02747}, 2022.

\bibitem[Liu et~al.(2025{\natexlab{a}})Liu, Zhang, Wang, Wei, Qiu, Zhao, Zhang, Ye, and Wan]{liu2025teacache}
Feng Liu, Shiwei Zhang, Xiaofeng Wang, Yujie Wei, Haonan Qiu, Yuzhong Zhao, Yingya Zhang, Qixiang Ye, and Fang Wan.
\newblock Timestep embedding tells: It's time to cache for video diffusion model.
\newblock In \emph{Proceedings of the Computer Vision and Pattern Recognition Conference}, pp.\  7353--7363, 2025{\natexlab{a}}.

\bibitem[Liu et~al.(2025{\natexlab{b}})Liu, Zou, Lyu, Chen, and Zhang]{liu2025taylor}
Jiacheng Liu, Chang Zou, Yuanhuiyi Lyu, Junjie Chen, and Linfeng Zhang.
\newblock From reusing to forecasting: Accelerating diffusion models with taylorseers, 2025{\natexlab{b}}.
\newblock URL \url{https://arxiv.org/abs/2503.06923}.

\bibitem[Liu et~al.(2025{\natexlab{c}})Liu, Zou, Lyu, Ren, Wang, Li, and Zhang]{liu2025speca}
Jiacheng Liu, Chang Zou, Yuanhuiyi Lyu, Fei Ren, Shaobo Wang, Kaixin Li, and Linfeng Zhang.
\newblock Speca: Accelerating diffusion transformers with speculative feature caching.
\newblock In \emph{Proceedings of the 33rd ACM International Conference on Multimedia}, pp.\  10024–10033. ACM, October 2025{\natexlab{c}}.
\newblock \doi{10.1145/3746027.3755331}.

\bibitem[Liu et~al.(2025{\natexlab{d}})Liu, Han, Xing, Yin, Wang, Cheng, Liao, Wang, Fu, Han, et~al.]{step1x2025liu}
Shiyu Liu, Yucheng Han, Peng Xing, Fukun Yin, Rui Wang, Wei Cheng, Jiaqi Liao, Yingming Wang, Honghao Fu, Chunrui Han, et~al.
\newblock Step1x-edit: A practical framework for general image editing.
\newblock \emph{arXiv preprint arXiv:2504.17761}, 2025{\natexlab{d}}.

\bibitem[Liu et~al.(2024)Liu, Tan, and Wang]{liu2024clear}
Songhua Liu, Zhenxiong Tan, and Xinchao Wang.
\newblock Clear: Conv-like linearization revs pre-trained diffusion transformers up.
\newblock \emph{arXiv preprint arXiv:2412.16112}, 2024.

\bibitem[Liu et~al.(2022)Liu, Gong, and Liu]{liu2022rectifiedflow}
Xingchao Liu, Chengyue Gong, and Qiang Liu.
\newblock Flow straight and fast: Learning to generate and transfer data with rectified flow.
\newblock \emph{arXiv preprint arXiv:2209.03003}, 2022.

\bibitem[Liu et~al.(2025{\natexlab{e}})Liu, Yang, Zhang, Zhang, Qiu, You, and Yang]{liu2025ras}
Ziming Liu, Yifan Yang, Chengruidong Zhang, Yiqi Zhang, Lili Qiu, Yang You, and Yuqing Yang.
\newblock Region-adaptive sampling for diffusion transformers.
\newblock \emph{arXiv preprint arXiv:2502.10389}, 2025{\natexlab{e}}.

\bibitem[Luo et~al.(2023)Luo, Tan, Huang, Li, and Zhao]{luo2023lcm}
Simian Luo, Yiqin Tan, Longbo Huang, Jian Li, and Hang Zhao.
\newblock Latent consistency models: Synthesizing high-resolution images with few-step inference.
\newblock \emph{arXiv preprint arXiv:2310.04378}, 2023.

\bibitem[Lyu et~al.(2022)Lyu, Xu, Yang, Lin, and Dai]{lyu2022earlystop}
Zhaoyang Lyu, Xudong Xu, Ceyuan Yang, Dahua Lin, and Bo~Dai.
\newblock Accelerating diffusion models via early stop of the diffusion process.
\newblock \emph{arXiv preprint arXiv:2205.12524}, 2022.

\bibitem[Ma et~al.(2024)Ma, Fang, and Wang]{ma2024deepcache}
Xinyin Ma, Gongfan Fang, and Xinchao Wang.
\newblock Deepcache: Accelerating diffusion models for free.
\newblock In \emph{Proceedings of the IEEE/CVF conference on computer vision and pattern recognition}, pp.\  15762--15772, 2024.

\bibitem[Pan et~al.(2023)Pan, Gherardi, Xie, and Huang]{pan2023effectiverealimageediting}
Zhihong Pan, Riccardo Gherardi, Xiufeng Xie, and Stephen Huang.
\newblock Effective real image editing with accelerated iterative diffusion inversion, 2023.

\bibitem[Peebles \& Xie(2023)Peebles and Xie]{peebles2023dit}
William Peebles and Saining Xie.
\newblock Scalable diffusion models with transformers.
\newblock In \emph{Proceedings of the IEEE/CVF international conference on computer vision}, pp.\  4195--4205, 2023.

\bibitem[Rombach et~al.(2022)Rombach, Blattmann, Lorenz, Esser, and Ommer]{rombach2022diff}
Robin Rombach, Andreas Blattmann, Dominik Lorenz, Patrick Esser, and Bj{\"o}rn Ommer.
\newblock High-resolution image synthesis with latent diffusion models.
\newblock In \emph{Proceedings of the IEEE/CVF conference on computer vision and pattern recognition}, pp.\  10684--10695, 2022.

\bibitem[Ronneberger et~al.(2015)Ronneberger, Fischer, and Brox]{unet2015ronneberger}
Olaf Ronneberger, Philipp Fischer, and Thomas Brox.
\newblock U-net: Convolutional networks for biomedical image segmentation.
\newblock In \emph{International Conference on Medical image computing and computer-assisted intervention}, pp.\  234--241. Springer, 2015.

\bibitem[Sauer et~al.(2024)Sauer, Lorenz, Blattmann, and Rombach]{sauer2024add}
Axel Sauer, Dominik Lorenz, Andreas Blattmann, and Robin Rombach.
\newblock Adversarial diffusion distillation.
\newblock In \emph{European Conference on Computer Vision}, pp.\  87--103. Springer, 2024.

\bibitem[Selvaraju et~al.(2024)Selvaraju, Ding, Chen, Zharkov, and Liang]{selvaraju2024fora}
Pratheba Selvaraju, Tianyu Ding, Tianyi Chen, Ilya Zharkov, and Luming Liang.
\newblock Fora: Fast-forward caching in diffusion transformer acceleration.
\newblock \emph{arXiv preprint arXiv:2407.01425}, 2024.

\bibitem[Shang et~al.(2023)Shang, Yuan, Xie, Wu, and Yan]{shang2023ptq4dm}
Yuzhang Shang, Zhihang Yuan, Bin Xie, Bingzhe Wu, and Yan Yan.
\newblock Post-training quantization on diffusion models.
\newblock In \emph{Proceedings of the IEEE/CVF conference on computer vision and pattern recognition}, pp.\  1972--1981, 2023.

\bibitem[Shen et~al.(2024)Shen, Chen, Ye, Xia, Chen, Bouganis, and Zhao]{shen2024mddit}
Mingzhu Shen, Pengtao Chen, Peng Ye, Guoxuan Xia, Tao Chen, Christos-Savvas Bouganis, and Yiren Zhao.
\newblock {MD}-dit: Step-aware mixture-of-depths for efficient diffusion transformers.
\newblock In \emph{Adaptive Foundation Models: Evolving AI for Personalized and Efficient Learning}, 2024.

\bibitem[Sun et~al.(2025)Sun, Tu, Ding, Jin, Liao, Liu, and Tao]{sun2025vorta}
Wenhao Sun, Rong-Cheng Tu, Yifu Ding, Zhao Jin, Jingyi Liao, Shunyu Liu, and Dacheng Tao.
\newblock Vorta: Efficient video diffusion via routing sparse attention, 2025.

\bibitem[Wang et~al.(2025)Wang, Pu, Qi, Guo, Ma, Huang, Chen, Li, and Shan]{wang2025tamingrectifiedflowinversion}
Jiangshan Wang, Junfu Pu, Zhongang Qi, Jiayi Guo, Yue Ma, Nisha Huang, Yuxin Chen, Xiu Li, and Ying Shan.
\newblock Taming rectified flow for inversion and editing, 2025.

\bibitem[Wang et~al.(2023)Wang, Zhang, Birsak, and Wonka]{wang2023instructedit}
Qian Wang, Biao Zhang, Michael Birsak, and Peter Wonka.
\newblock Instructedit: Improving automatic masks for diffusion-based image editing with user instructions.
\newblock \emph{arXiv preprint arXiv:2305.18047}, 2023.

\bibitem[Wang \& Bovik(2002)Wang and Bovik]{wang2002ssim}
Zhou Wang and Alan~C Bovik.
\newblock A universal image quality index.
\newblock \emph{IEEE signal processing letters}, 9\penalty0 (3):\penalty0 81--84, 2002.

\bibitem[Wu et~al.(2025)Wu, Li, Zhou, Lin, Gao, Yan, Yin, Bai, Xu, Chen, et~al.]{qwenimage2025wu}
Chenfei Wu, Jiahao Li, Jingren Zhou, Junyang Lin, Kaiyuan Gao, Kun Yan, Sheng-ming Yin, Shuai Bai, Xiao Xu, Yilei Chen, et~al.
\newblock Qwen-image technical report.
\newblock \emph{arXiv preprint arXiv:2508.02324}, 2025.

\bibitem[Xi et~al.(2025)Xi, Yang, Zhao, Xu, Li, Li, Lin, Cai, Zhang, Li, Chen, Stoica, Keutzer, and Han]{xi2025svg}
Haocheng Xi, Shuo Yang, Yilong Zhao, Chenfeng Xu, Muyang Li, Xiuyu Li, Yujun Lin, Han Cai, Jintao Zhang, Dacheng Li, Jianfei Chen, Ion Stoica, Kurt Keutzer, and Song Han.
\newblock Sparse videogen: Accelerating video diffusion transformers with spatial-temporal sparsity, 2025.

\bibitem[Yan et~al.(2025)Yan, Ma, Zou, Chen, Chen, and Zhang]{yan2025eedit}
Zexuan Yan, Yue Ma, Chang Zou, Wenteng Chen, Qifeng Chen, and Linfeng Zhang.
\newblock Eedit: Rethinking the spatial and temporal redundancy for efficient image editing, 2025.

\bibitem[Yuan et~al.(2024)Yuan, Zhang, Lu, Ning, Zhang, Zhao, Yan, Dai, and Wang]{yuan2025ditfastattn}
Zhihang Yuan, Hanling Zhang, Pu~Lu, Xuefei Ning, Linfeng Zhang, Tianchen Zhao, Shengen Yan, Guohao Dai, and Yu~Wang.
\newblock Ditfastattn: Attention compression for diffusion transformer models.
\newblock In A.~Globerson, L.~Mackey, D.~Belgrave, A.~Fan, U.~Paquet, J.~Tomczak, and C.~Zhang (eds.), \emph{Advances in Neural Information Processing Systems}, volume~37, pp.\  1196--1219. Curran Associates, Inc., 2024.

\bibitem[Zhang et~al.(2025)Zhang, Su, Yuan, Chen, Shen, Fan, Yan, Dai, and Wang]{zhang2025ditfastattnv2}
Hanling Zhang, Rundong Su, Zhihang Yuan, Pengtao Chen, Mingzhu Shen, Yibo Fan, Shengen Yan, Guohao Dai, and Yu~Wang.
\newblock Ditfastattnv2: Head-wise attention compression for multi-modality diffusion transformers.
\newblock In \emph{Proceedings of the IEEE/CVF International Conference on Computer Vision (ICCV)}, pp.\  16399--16409, October 2025.

\bibitem[Zhang et~al.(2023{\natexlab{a}})Zhang, Mo, Chen, Sun, and Su]{zhang2023magicbrush}
Kai Zhang, Lingbo Mo, Wenhu Chen, Huan Sun, and Yu~Su.
\newblock Magicbrush: A manually annotated dataset for instruction-guided image editing.
\newblock \emph{Advances in Neural Information Processing Systems}, 36:\penalty0 31428--31449, 2023{\natexlab{a}}.

\bibitem[Zhang et~al.(2023{\natexlab{b}})Zhang, Rao, and Agrawala]{controlnet2023zhang}
Lvmin Zhang, Anyi Rao, and Maneesh Agrawala.
\newblock Adding conditional control to text-to-image diffusion models.
\newblock In \emph{Proceedings of the IEEE/CVF international conference on computer vision}, pp.\  3836--3847, 2023{\natexlab{b}}.

\bibitem[Zhang et~al.(2018)Zhang, Isola, Efros, Shechtman, and Wang]{zhang2018LPIPS}
Richard Zhang, Phillip Isola, Alexei~A Efros, Eli Shechtman, and Oliver Wang.
\newblock The unreasonable effectiveness of deep features as a perceptual metric.
\newblock In \emph{Proceedings of the IEEE conference on computer vision and pattern recognition}, pp.\  586--595, 2018.

\bibitem[Zhao et~al.(2025)Zhao, Chen, Yu, Wen, Tan, and Chen]{zhao2025pioneering}
Maosen Zhao, Pengtao Chen, Chong Yu, Yan Wen, Xudong Tan, and Tao Chen.
\newblock Pioneering 4-bit fp quantization for diffusion models: Mixup-sign quantization and timestep-aware fine-tuning, 2025.

\bibitem[Zhao et~al.(2024)Zhao, Jin, Wang, and You]{zhao2024psnr}
Xuanlei Zhao, Xiaolong Jin, Kai Wang, and Yang You.
\newblock Real-time video generation with pyramid attention broadcast.
\newblock \emph{arXiv preprint arXiv:2408.12588}, 2024.

\bibitem[Zhou et~al.(2025)Zhou, Liang, Chen, Feng, Chen, Lin, Ding, Tan, Zhao, and Bai]{zhou2025easycache}
Xin Zhou, Dingkang Liang, Kaijin Chen, Tianrui Feng, Xiwu Chen, Hongkai Lin, Yikang Ding, Feiyang Tan, Hengshuang Zhao, and Xiang Bai.
\newblock Less is enough: Training-free video diffusion acceleration via runtime-adaptive caching, 2025.

\bibitem[Zou et~al.(2024{\natexlab{a}})Zou, Liu, Liu, Huang, and Zhang]{zou2024toca}
Chang Zou, Xuyang Liu, Ting Liu, Siteng Huang, and Linfeng Zhang.
\newblock Accelerating diffusion transformers with token-wise feature caching.
\newblock \emph{arXiv preprint arXiv:2410.05317}, 2024{\natexlab{a}}.

\bibitem[Zou et~al.(2024{\natexlab{b}})Zou, Zhang, Guo, Xu, He, Hu, and Zhang]{zou2024duca}
Chang Zou, Evelyn Zhang, Runlin Guo, Haohang Xu, Conghui He, Xuming Hu, and Linfeng Zhang.
\newblock Accelerating diffusion transformers with dual feature caching.
\newblock \emph{arXiv preprint arXiv:2412.18911}, 2024{\natexlab{b}}.

\end{thebibliography}
\bibliographystyle{iclr2026_conference}

\clearpage
\appendix
\centering \textbf{\large RegionE: Adaptive Region-Aware Generation for Efficient Image Editing}

\vspace{5pt}
\centering {\large Supplementary Material}

\justify

\vspace{10pt}

We organize the supplementary material as follows:


\begin{itemize}
    \item \cref{pseudocode}: Pseudocode of RegionE
    \item \cref{sec:cache}: Analysis of Adaptive Velocity Decay Cache
    \item \cref{sec:visualization}: Per-Task Visualization Results in the Benchmark
    \item \cref{sec:quantative}: Per-Task Quantitative Results in the Benchmark
\end{itemize}

\clearpage
\section{Pseudocode of RegionE}
\label{pseudocode}
\begin{algorithm}[H]
\caption{RegionE: Adaptive Region-Aware Generation for Efficient Image Editing}
\label{algorithm}
\begin{algorithmic}[1]
    \REQUIRE Diffusion transformer $\Phi(\cdot)$, sampling step $T$, insturction image $\bm X^I$, text tokens $\bm X^P$, random noise $\bm X_T$, total steps in stabilization stage $t^{st}$, total steps in smooth stage $t^{sm}$, threshold of adaptive region partition $\eta$, threshold of adaptive velocity decay cache $\delta$, sorted forced steps list $t_{f} \_list$.
    \STATE // \textbf{Initialization}
    \STATE RIKVCache $\mathcal{C_{KV}}$ = None, RIKVCache flag $f=(False,False)$; AVDCache $\mathcal{C_A}$=None;
    \STATE Accumulative Error $e=0$; $t_{f} \_list.insert(0,T-t^{st})$; $t_{f} \_list.insert(-1,t^{sm}-1)$;
    \STATE // \textbf{Stabilization Stage}
    \FOR{$i \gets T$ to $T-t^{st}$}
        \IF{$i==T-t^{st}$}
            \STATE $f[0]=True$  \COMMENT{first dimension represents storing, second dimension represents retrieving}
        \ENDIF
        \STATE $\bm v_{t_i},\mathcal{C_{KV}} = \Phi([\bm X^P, \bm X_{t_i}, \bm X^I],\mathcal{C_{KV}},f)$
        \STATE $\bm X_{t_{i-1}}=\bm X_{t_{i}} - (t_i-t_{i-1})\cdot \bm  v_{t_i}$
    \ENDFOR
    
    \STATE // \textbf{Region-Aware Generation Stage}
    \STATE \COMMENT{Adaptive Region Partition}
    \STATE $\hat{\bm{X}_0}=\bm X_{t_{T-t^{st}}}-\bm v_{T-t^{st}+1}\cdot t_{T-t^{st}} $
    \STATE $E_{index},U_{index}=Erosion\_\&\_Dilate(cos(\hat{\bm{X}_0},\bm X^I)>\eta)$
    \STATE \COMMENT{Region-Aware Generation}
    \FOR{$i \gets 0$ to $len(t_{f} \_list) - 2$}
        \STATE $prev = t_{f} \_list[i]$; $next = t_{f} \_list[i+1]$
        \STATE $\bm X^E_{t_{prev}} = \bm X_{t_{prev}}[E_{index}];\bm X^U_{t_{prev}} = \bm X_{t_{prev}}[U_{index}]$
        \STATE $\hat{\bm{X}}^U_{t_{next+1}}=\bm X_{t_{prev}}^U-\bm v^U_{t_{prev}+1}\cdot (t_{prev}-t_{next+1})$ \COMMENT{one-step estimation for unedited region}
        \STATE $f[0]=False$, $f[1]=True$    \COMMENT{iteritive denoising for edited region}
        \FOR{$j \gets prev$ to $next+1$}
            \STATE \COMMENT{Adaptive Velocity Decay Cache}
            \STATE Calculate $e$ according to Eq.\ref{eq:accumulate}
            \IF{$e > \delta$}
                \STATE $\bm v^E_{t_j},\mathcal{C_{KV}} = \Phi([\bm X^P, \bm X^E_{t_j}, \bm X^I],\mathcal{C_{KV}},f)$
                \STATE $\mathcal{C_A} =\bm v^E_{t_j}$
                \STATE $\bm X^E_{t_{j-1}}=\bm X^E_{t_{j}} - (t_j-t_{j-1})\cdot \bm  v^E_{t_j}$
            \ELSE
                \STATE $\bm v^E_{t_j}=\mathcal{C_A}*$decay factor according to Eq.\ref{eq:decay}
            \ENDIF
        \ENDFOR
        \STATE $\bm X_{t_{next+1}}=gather(\bm X^U_{t_{next}}, \bm X^E_{t_{next+1}})$
        \STATE $f[0]=True, f[1]=False$
        \STATE $\bm v_{t_{next+1}},\mathcal{C_{KV}} = \Phi([\bm X^P, \bm X_{t_{next+1}}, \bm X^I],\mathcal{C_{KV}},f)$
        \STATE $\bm X_{t_{next}}=\bm X_{t_{next+1}} - (t_{next}-t_{next+1})\cdot \bm  v_{t_{next+1}}$
    \ENDFOR
    \STATE // \textbf{Smooth Stage}
    \STATE $f[0]=False, f[1]=False$
    \FOR{$i \gets t^{sm}-1$ to $1$}
        \STATE $\bm v_{t_i},\mathcal{C_{KV}} = \Phi(\bm X_{t_i},\mathcal{C_{KV}},f)$
        \STATE $\bm X_{t_{i-1}}=\bm X_{t_{i}} - (t_i-t_{i-1})\cdot \bm  v_{t_i}$
    \ENDFOR
    \ENSURE Target image after editing $\bm X_0$
\end{algorithmic}
\end{algorithm}

\clearpage
\section{Analysis of Adaptive Velocity Decay Cache}
\label{sec:cache}
\begin{figure*}[h]
    \centering
    \includegraphics[width=0.99 \linewidth]{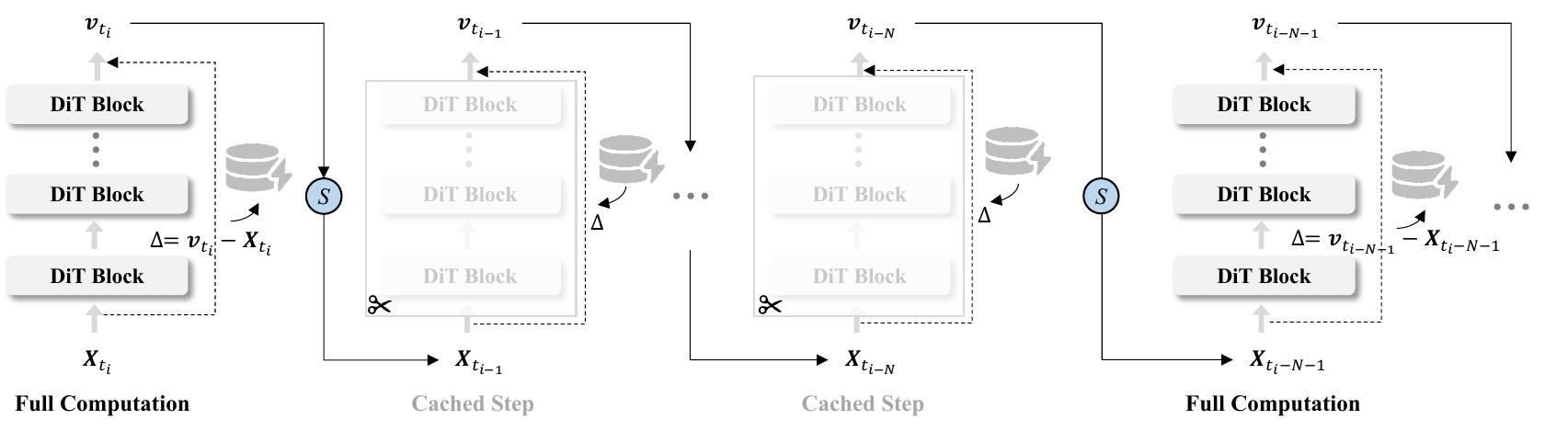}
    \caption{Pipeline Based on Residual Cache.}
    \label{fig:deltacache}
\end{figure*}

In current research on diffusion model caching, many studies focus on residual-based caches~\citep{chen2024deltadit, liu2025teacache, zhou2025easycache, bu2025dicacheletdiffusionmodel}, which store the $\Delta$ shown in Figure~\ref{fig:deltacache}. Based on the sampling formula in Equation~\ref{eq:inference} and the definition of caching, we can derive the following expression:
\begin{equation}
\left\{
\begin{aligned}
\bm X_{t_{i-1}} &= \bm X_{t_i} - (t_i-t_{i-1})\cdot \bm v_{t_i} \\
\Delta &= \bm v_{t_i} - \bm X_{t_i} \\
\bm v_{t_{i-1}} &= \bm X_{t_{i-1}} + \Delta
\end{aligned}
\right..
\end{equation}
It can be solved as:
\begin{equation}
\bm v_{t_{i-1}}=[1-(t_i-t_{i-1})]\cdot \bm v_{t_i}.
\end{equation}
Similarly, for the timestep $t_{i-2}$, we have:
\begin{equation}
\bm v_{t_{i-2}}=[1-(t_{i-1}-t_{i-2})]\cdot \bm v_{t_{i-1}}.
\end{equation}
Therefore, if we perform $N$ steps of residual caching, as illustrated in Figure~\ref{fig:deltacache}, we can obtain:
\begin{equation}
\begin{aligned}
\bm v_{t_{i-N}} &=\prod_{m=1}^{N} [1-(t_{i-m+1}-t_{i-m})]\cdot \bm v_{t_i} \\
&= \underbrace{\prod_{m=1}^{N} [1-\Delta t_{{i-m+1},{i-m}}]}_{\text{Determined by Solver}}\cdot \bm v_{t_i}. \\
\label{eq:deltacache}
\end{aligned}
\end{equation}
This further indicates that the current residual cache and the velocity cache are equivalent. Since $\Delta t_{{i-m+1},{i-m}}$ is a minimal value approaching zero, the coefficient before $\bm v_{t_i}$ is less than one. Therefore, it can be seen that the current residual cache is essentially a decayed form of the velocity cache. Furthermore, we observe that the solver determines the decay coefficient in Equation~\ref{eq:deltacache}. However, as shown in Figure~\ref{fig:proof}d, the decay of velocity exhibits a timestep-dependent behavior. To account for this, we introduce an external timestep correction coefficient $\gamma_{t_{i}}$. Notably, the AVDCache proposed in this paper reduces to Equation~\ref{eq:deltacache} when the correction coefficient $\gamma_{t_{i}}$ equals 1.



\clearpage
\section{Per-Task Visualization Results in the Benchmark}
\label{sec:visualization}
\begin{figure*}[t]
    \centering
    \includegraphics[width=0.95 \linewidth]{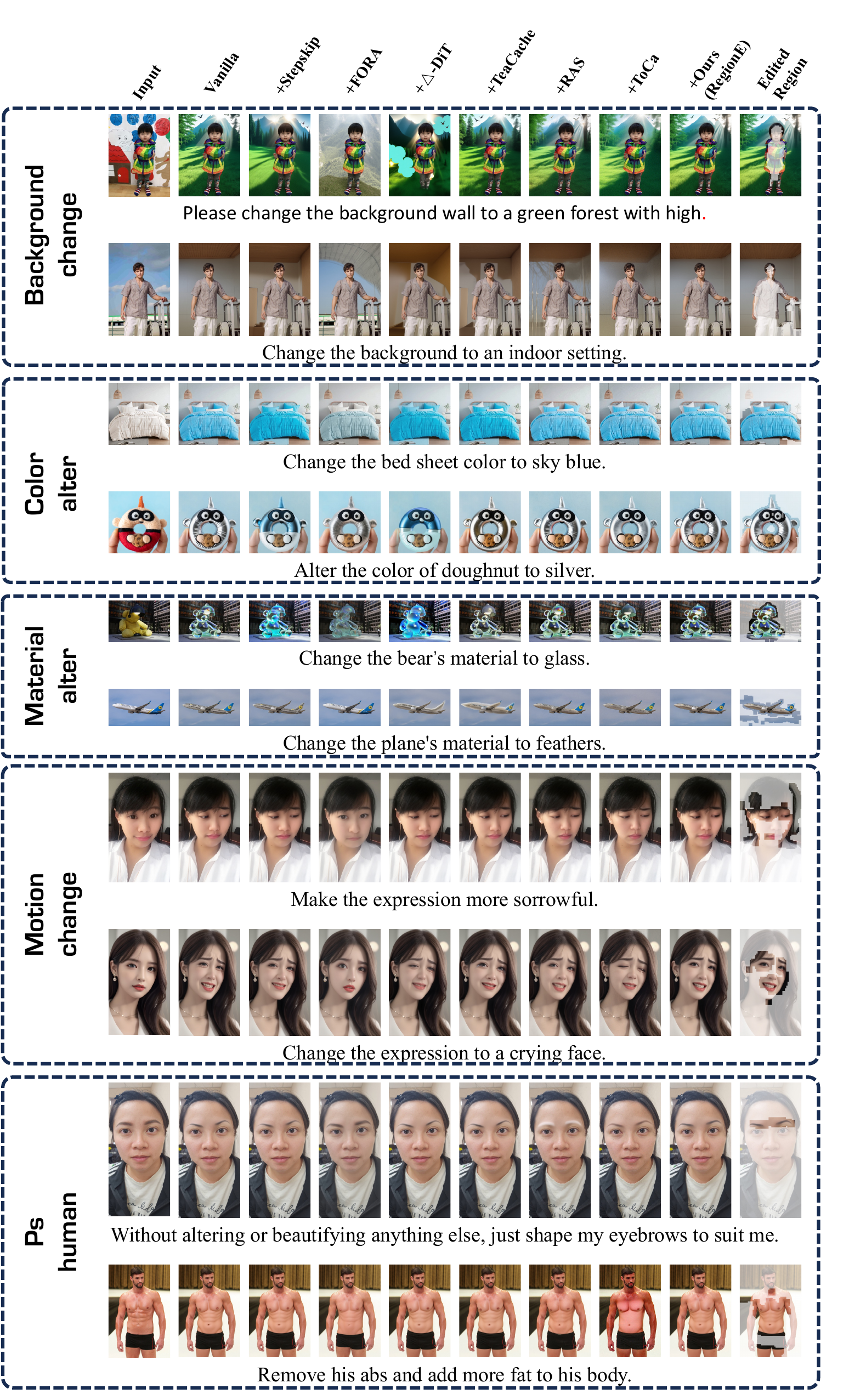}
    \caption{Examples of edited images by RegionE and baseline on Step1X-Edit-v1p1.}
    \label{fig:visu_step1_2}
\end{figure*}

\begin{figure*}[t]
    \centering
    \includegraphics[width=0.99 \linewidth]{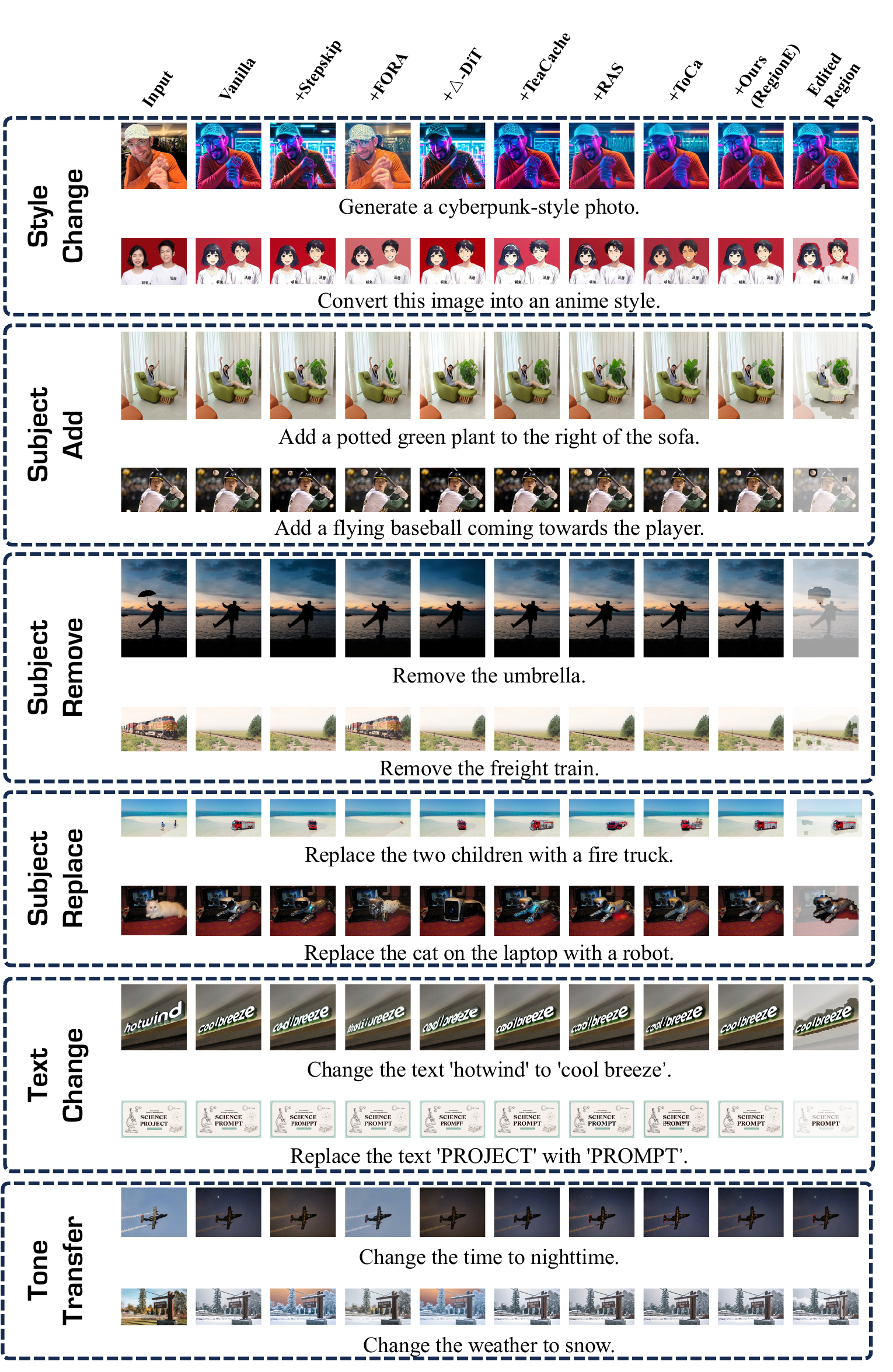}
    \caption{Examples of edited images by RegionE and baseline on Step1X-Edit-v1p1.}
    \label{fig:visu_step2_2}
\end{figure*}

\begin{figure*}[t]
    \centering
    \includegraphics[width=0.99 \linewidth]{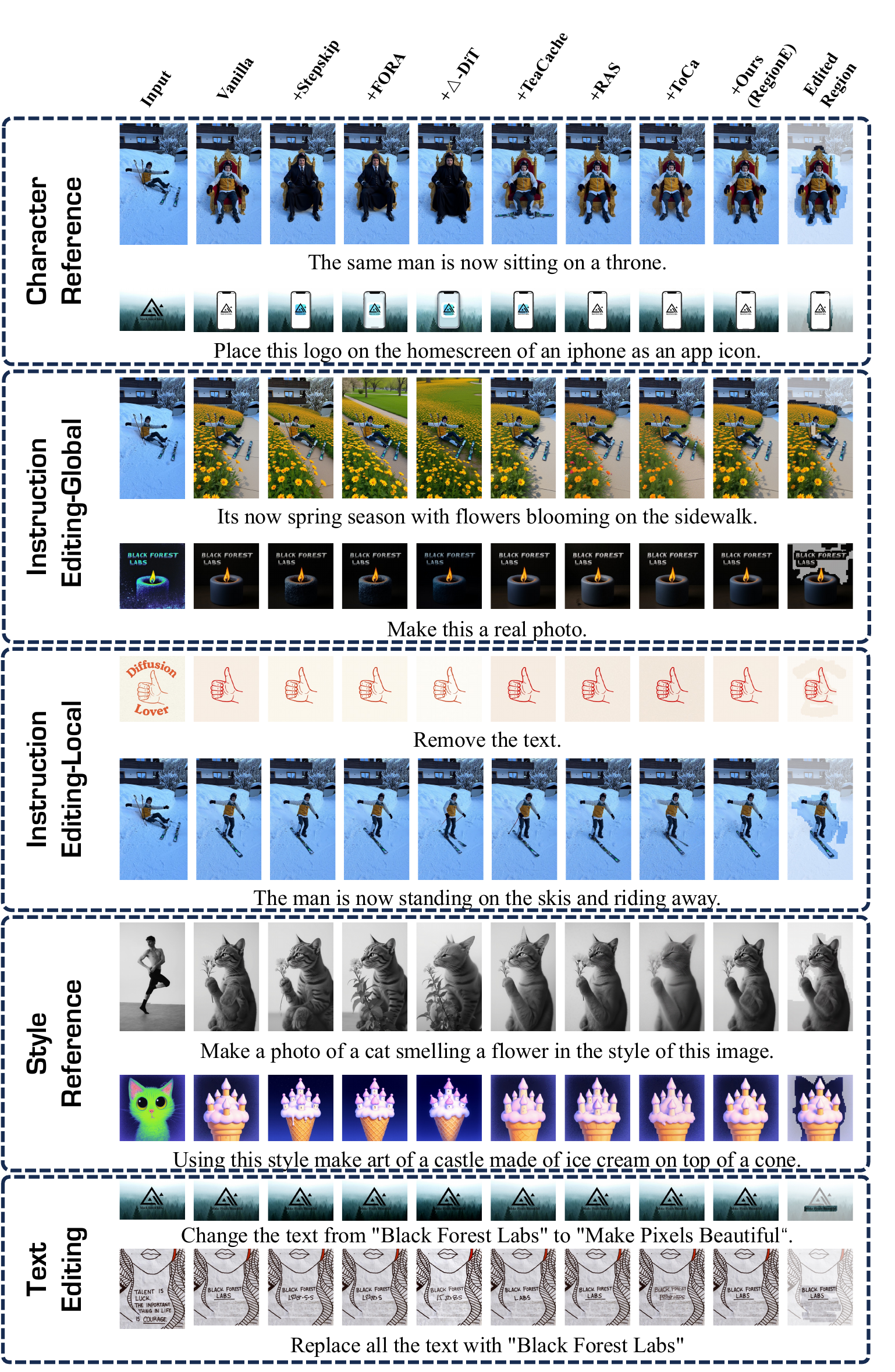}
    \caption{Examples of edited images by RegionE and baseline on FLUX.1 Kontext.}
    \label{fig:visu_flux1_1}
\end{figure*}

\begin{figure*}[t]
    \centering
    \includegraphics[width=0.95 \linewidth]{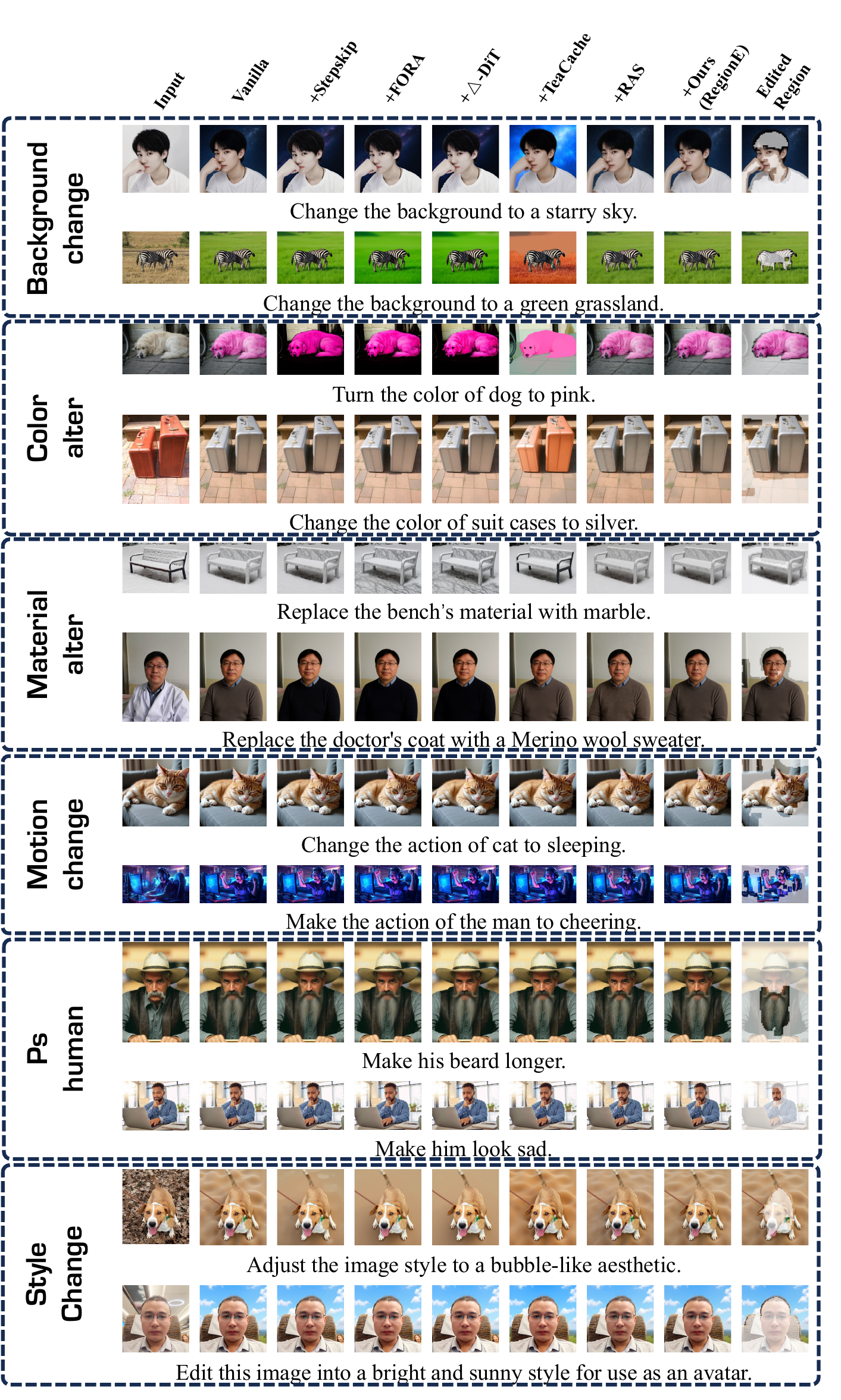}
    \caption{Examples of edited images by RegionE and baseline on Qwen-Image-Edit.}
    \label{fig:visu_qwen1_2}
\end{figure*}

\begin{figure*}[t]
    \centering
    \includegraphics[width=0.97 \linewidth]{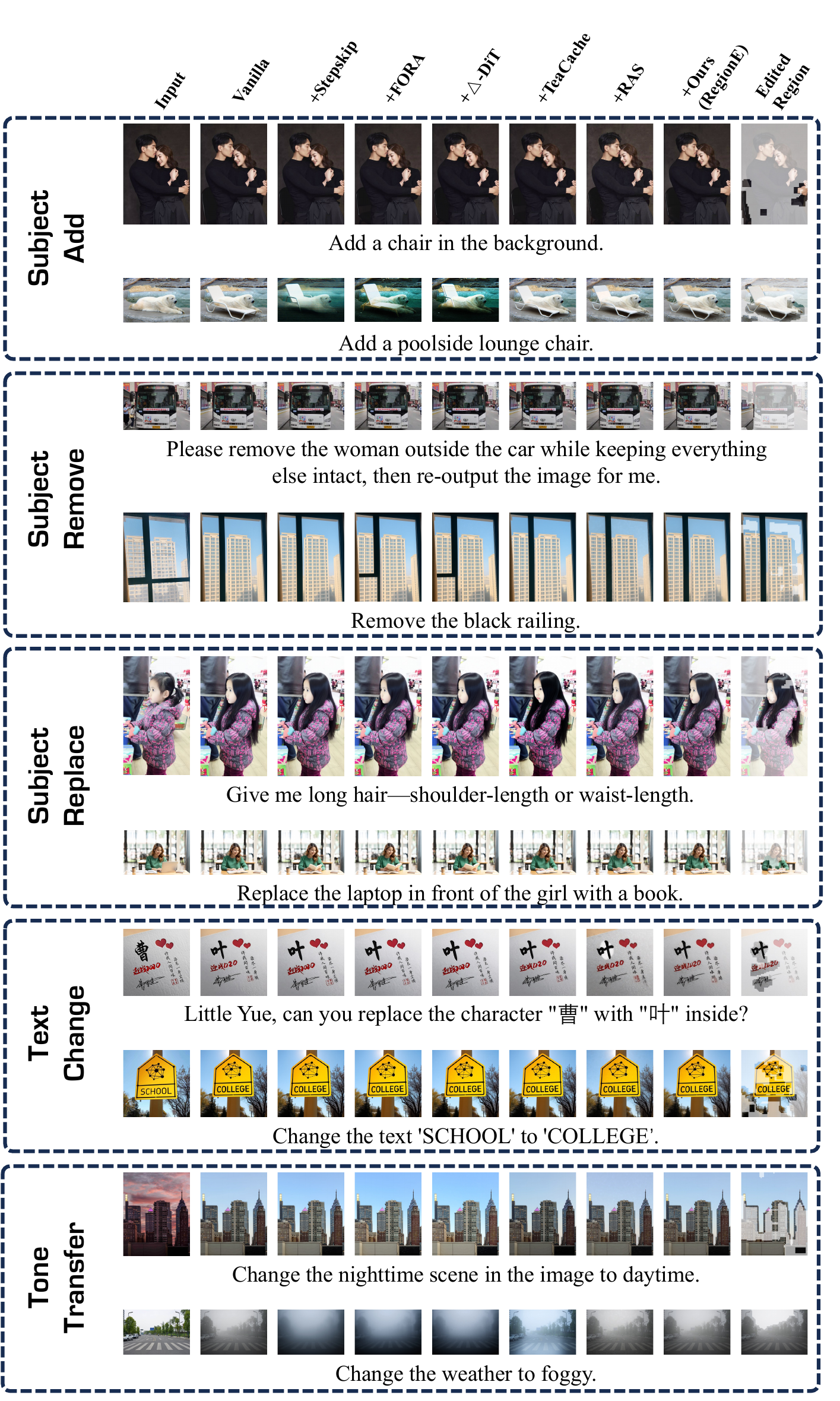}
    \caption{Examples of edited images by RegionE and baseline on Qwen-Image-Edit.}
    \label{fig:visu_qwen2_2}
\end{figure*}

Due to space limitations, we put the visualization results of some tasks in the manuscript. Here, we provide a visual comparison of additional tasks and models. Figure~\ref{fig:visu_step1_2} and Figure~\ref{fig:visu_step2_2} show the visualization results of 11 tasks on Step1X-Edit. Figure~\ref{fig:visu_qwen1_2} and Figure~\ref{fig:visu_qwen2_2} show the visualization results of 11 tasks on Qwen-Image-Edit. Figure~\ref{fig:visu_flux1_1} show the visualization results of 5 tasks on FLUX.1 Kontext.

\clearpage
\section{Per-Task Quantitative Results in the Benchmark}
\label{sec:quantative}
In this section, we present the performance of RegionE and the baseline methods on each task in the benchmark. Table~\ref{tab:11_motion_change}-Table~\ref{tab:11_background_change} show the performance on the 11 tasks: motion-change, ps-human, color-alter, material-alter, subject-add, subject-remove, style-change, tone-transfer, subject-replace, text-change, and background-change. Table~\ref{tab:5_CR}-Table\ref{tab:5_TE} show the performance on the five tasks: Character Reference, Style Reference, Text Editing, Instruction Editing-Global, and Instruction Editing-Local.
\begin{table}[h]
    \caption{Comparison of RegionE and other baselines on the motion-change task of GEdit-Bench, evaluated in terms of quality and efficiency.}
    \resizebox{\linewidth}{!}{%
    \begin{tabular}{l|ccc|ccc|cc}
        \toprule[1.5pt]
         \multirow{2}{*}{\textbf{Model}}&  \multicolumn{3}{c|}{\textbf{Against Vanilla}}& \multicolumn{3}{c|}{\textbf{GPT-4o Score}}&  \multicolumn{2}{c}{\textbf{Efficiency}}\\
 & \textbf{PSNR$\uparrow$ }& \textbf{SSIM $\uparrow$ }& \textbf{LPIPS $\downarrow$ }& \textbf{G-SC $\uparrow$ }&\textbf{G-PQ $\uparrow$ }& \textbf{G-O $\uparrow$ }& \textbf{Latency (s) $\downarrow$ }&\textbf{Speedup $\uparrow$ }\\
        \midrule
         \textbf{Step1X-Edit}~\citep{step1x2025liu}&  -&  -&  -& 4.350 &7.950 &  4.444 &  27.950 &1.000 
\\
         \; \textbf{+ Stepskip}&  25.887 &  0.902 &  0.093 & 4.350 &8.100 &  4.562 &  12.306 &2.271 
\\
         \; \textbf{+ FORA}~\citep{selvaraju2024fora}&  20.935 &  0.818 &  0.189 & 2.175 &7.575 &  2.385 &  14.339 &1.949 
\\
         \; \textbf{+ $\Delta$-DiT}~\citep{chen2024deltadit}&  24.549 &  0.876 &  0.121 & 4.350 &7.975 &  4.445 &  12.730 &2.196 
\\
         \; \textbf{+ TeaCache}~\citep{liu2025teacache}&  26.926 &  0.925 &  0.068 & 4.475 &8.050 &  4.524 &  11.218 &2.492 
\\
         \; \textbf{+ RAS}~\citep{liu2025ras}&  25.888 &  0.889 &  0.109 & 4.025 &7.375 &  4.012 &  15.253 &1.832 
\\
         \; \textbf{+ ToCa}~\citep{zou2024toca}& 24.428 & 0.843 & 0.165 & 3.775 &6.975 & 3.578 & 22.225 &1.258 
\\
        
        \rowcolor[HTML]{EFEFEF}
         \; \textbf{+ Ours (RegionE)}& 29.633 & 0.937 & 0.053 & 4.625 &7.775 & 4.763 & 10.739 &2.603 
\\
         \midrule
         \textbf{Qwen-Image-Edit}~\citep{qwenimage2025wu}& -& -& -& 4.850 &8.550 & 5.112 & 32.140 &1.000 
\\
         \; \textbf{+ Stepskip}
& 27.791 & 0.905 & 0.066 & 4.725 &8.625 & 5.029 & 17.566 &1.830 
\\
         \; \textbf{+ FORA}~\citep{selvaraju2024fora}
& 26.744 & 0.889 & 0.079 & 4.825 &8.325 & 4.995 & 17.827 &1.803 
\\
         \; \textbf{+ $\Delta$-DiT}~\citep{chen2024deltadit}
& 25.756 & 0.848 & 0.095 & 4.675 &8.575 & 4.921 & 17.481 &1.839 
\\
         \; \textbf{+ TeaCache}~\citep{liu2025teacache}
& 26.776 & 0.911 & 0.070 & 5.025 &8.500 & 5.251 & 16.389 &1.961 
\\
         \; \textbf{+ RAS}~\citep{liu2025ras}
& 26.585 & 0.882 & 0.096 & 5.000 &8.625 & 5.262 & 22.300 &1.441 
\\
         \; \textbf{+ ToCa}~\citep{zou2024toca}
& OOM& OOM& OOM& OOM&OOM& OOM& OOM&OOM\\
        
         \rowcolor[HTML]{EFEFEF}
         \; \textbf{+ Ours (RegionE)}& 29.416 & 0.932 & 0.057 & 4.825 &8.550 & 5.164 & 15.695 &2.048 
\\
         \bottomrule[1.5pt]
    \end{tabular}
    }
    \label{tab:11_motion_change}
\end{table}
\begin{table}[h]
    \caption{Comparison of RegionE and other baselines on the ps-human task of GEdit-Bench, evaluated in terms of quality and efficiency.}
    \resizebox{\linewidth}{!}{%
    \begin{tabular}{l|ccc|ccc|cc}
        \toprule[1.5pt]
         \multirow{2}{*}{\textbf{Model}}&  \multicolumn{3}{c|}{\textbf{Against Vanilla}}& \multicolumn{3}{c|}{\textbf{GPT-4o Score}}&  \multicolumn{2}{c}{\textbf{Efficiency}}\\
 & \textbf{PSNR$\uparrow$ }& \textbf{SSIM $\uparrow$ }& \textbf{LPIPS $\downarrow$ }& \textbf{G-SC $\uparrow$ }&\textbf{G-PQ $\uparrow$ }& \textbf{G-O $\uparrow$ }& \textbf{Latency (s) $\downarrow$ }&\textbf{Speedup $\uparrow$ }\\
        \midrule
         \textbf{Step1X-Edit}~\citep{step1x2025liu}&  -&  -&  -& 4.614 &8.086 &  4.649 &  27.927 &1.000 
\\
         \; \textbf{+ Stepskip}&  29.220 &  0.916 &  0.069 & 4.600 &8.086 &  4.728 &  12.296 &2.271 
\\
         \; \textbf{+ FORA}~\citep{selvaraju2024fora}&  23.596 &  0.863 &  0.142 & 3.414 &8.529 &  3.920 &  14.323 &1.950 
\\
         \; \textbf{+ $\Delta$-DiT}~\citep{chen2024deltadit}&  26.348 &  0.884 &  0.099 & 4.800 &8.086 &  4.893 &  12.728 &2.194 
\\
         \; \textbf{+ TeaCache}~\citep{liu2025teacache}&  31.428 &  0.942 &  0.047 & 5.114 &7.929 &  5.191 &  11.208 &2.492 
\\
         \; \textbf{+ RAS}~\citep{liu2025ras}&  29.077 &  0.921 &  0.072 & 4.400 &7.886 &  4.486 &  15.237 &1.833 
\\
         \; \textbf{+ ToCa}~\citep{zou2024toca}& 26.716 & 0.878 & 0.125 & 4.786 &7.914 & 4.838 & 22.073 &1.265 
\\
        
        \rowcolor[HTML]{EFEFEF}
         \; \textbf{+ Ours (RegionE)}& 32.985 & 0.957 & 0.037 & 4.629 &8.114 & 4.731 & 10.813 &2.583 
\\
         \midrule
         \textbf{Qwen-Image-Edit}~\citep{qwenimage2025wu}& -& -& -& 5.814 &8.500 & 5.972 & 32.100 &1.000 
\\
         \; \textbf{+ Stepskip}
& 32.080 & 0.936 & 0.040 & 5.757 &8.414 & 5.904 & 17.553 &1.829 
\\
         \; \textbf{+ FORA}~\citep{selvaraju2024fora}
& 30.120 & 0.920 & 0.049 & 5.700 &8.443 & 5.933 & 17.816 &1.802 
\\
         \; \textbf{+ $\Delta$-DiT}~\citep{chen2024deltadit}
& 28.323 & 0.887 & 0.062 & 5.743 &8.500 & 5.911 & 17.462 &1.838 
\\
         \; \textbf{+ TeaCache}~\citep{liu2025teacache}
& 32.347 & 0.948 & 0.038 & 5.714 &8.400 & 5.833 & 16.360 &1.962 
\\
         \; \textbf{+ RAS}~\citep{liu2025ras}
& 29.857 & 0.917 & 0.061 & 5.843 &8.271 & 5.884 & 22.340 &1.437 
\\
         \; \textbf{+ ToCa}~\citep{zou2024toca}
& OOM& OOM& OOM& OOM&OOM& OOM& OOM&OOM\\
        
         \rowcolor[HTML]{EFEFEF}
         \; \textbf{+ Ours (RegionE)}& 33.550 & 0.963 & 0.029 & 6.086 &8.486 & 6.227 & 15.473 &2.075 
\\
         \bottomrule[1.5pt]
    \end{tabular}
    }
    \label{tab:11_ps_human}
\end{table}
\begin{table}
    \caption{Comparison of RegionE and other baselines on the color-alter task of GEdit-Bench, evaluated in terms of quality and efficiency.}
    \resizebox{\linewidth}{!}{%
    \begin{tabular}{l|ccc|ccc|cc}
        \toprule[1.5pt]
         \multirow{2}{*}{\textbf{Model}}&  \multicolumn{3}{c|}{\textbf{Against Vanilla}}& \multicolumn{3}{c|}{\textbf{GPT-4o Score}}&  \multicolumn{2}{c}{\textbf{Efficiency}}\\
 & \textbf{PSNR$\uparrow$ }& \textbf{SSIM $\uparrow$ }& \textbf{LPIPS $\downarrow$ }& \textbf{G-SC $\uparrow$ }&\textbf{G-PQ $\uparrow$ }& \textbf{G-O $\uparrow$ }& \textbf{Latency (s) $\downarrow$ }&\textbf{Speedup $\uparrow$ }\\
        \midrule
         \textbf{Step1X-Edit}~\citep{step1x2025liu}&  -&  -&  -& 8.750 &6.875 &  7.395 &  28.019 &1.000 
\\
         \; \textbf{+ Stepskip}&  27.291 &  0.919 &  0.080 & 8.325 &6.975 &  7.349 &  12.330 &2.273 
\\
         \; \textbf{+ FORA}~\citep{selvaraju2024fora}&  21.871 &  0.838 &  0.132 & 8.800 &7.525 &  7.889 &  14.356 &1.952 
\\
         \; \textbf{+ $\Delta$-DiT}~\citep{chen2024deltadit}&  24.942 &  0.901 &  0.107 & 8.075 &6.600 &  6.968 &  12.770 &2.194 
\\
         \; \textbf{+ TeaCache}~\citep{liu2025teacache}&  28.084 &  0.938 &  0.050 & 8.525 &6.950 &  7.345 &  11.242 &2.492 
\\
         \; \textbf{+ RAS}~\citep{liu2025ras}&  28.800 &  0.909 &  0.069 & 8.700 &6.925 &  7.432 &  15.274 &1.834 
\\
         \; \textbf{+ ToCa}~\citep{zou2024toca}& 25.917 & 0.864 & 0.118 & 8.600 &6.725 & 7.232 & 21.996 &1.274 
\\
        
        \rowcolor[HTML]{EFEFEF}
         \; \textbf{+ Ours (RegionE)}& 32.739 & 0.956 & 0.032 & 8.850 &7.250 & 7.747 & 11.188 &2.504 
\\
         \midrule
         \textbf{Qwen-Image-Edit}~\citep{qwenimage2025wu}& inf& 1.000 & 0.000 & 9.250 &7.525 & 8.170 & 32.082 &1.000 
\\
         \; \textbf{+ Stepskip}
& 29.795 & 0.896 & 0.064 & 9.050 &7.450 & 8.084 & 17.527 &1.830 
\\
         \; \textbf{+ FORA}~\citep{selvaraju2024fora}
& 28.035 & 0.879 & 0.078 & 8.875 &7.350 & 7.872 & 17.795 &1.803 
\\
         \; \textbf{+ $\Delta$-DiT}~\citep{chen2024deltadit}
& 25.892 & 0.835 & 0.094 & 9.025 &7.375 & 8.021 & 17.479 &1.835 
\\
         \; \textbf{+ TeaCache}~\citep{liu2025teacache}
& 30.757 & 0.922 & 0.057 & 8.775 &7.250 & 7.840 & 16.566 &1.937 
\\
         \; \textbf{+ RAS}~\citep{liu2025ras}
& 29.132 & 0.909 & 0.060 & 9.150 &7.050 & 7.860 & 22.356 &1.435 
\\
         \; \textbf{+ ToCa}~\citep{zou2024toca}
& OOM& OOM& OOM& OOM&OOM& OOM& OOM&OOM\\
        
         \rowcolor[HTML]{EFEFEF}
         \; \textbf{+ Ours (RegionE)}& 33.144 & 0.951 & 0.032 & 9.225 &7.475 & 8.172 & 15.527 &2.066 
\\
         \bottomrule[1.5pt]
    \end{tabular}
    }
    \label{tab:11_color_alter}
\end{table}
\begin{table}
    \caption{Comparison of RegionE and other baselines on the material-alter task of GEdit-Bench, evaluated in terms of quality and efficiency.}
    \resizebox{\linewidth}{!}{%
    \begin{tabular}{l|ccc|ccc|cc}
        \toprule[1.5pt]
         \multirow{2}{*}{\textbf{Model}}&  \multicolumn{3}{c|}{\textbf{Against Vanilla}}& \multicolumn{3}{c|}{\textbf{GPT-4o Score}}&  \multicolumn{2}{c}{\textbf{Efficiency}}\\
 & \textbf{PSNR$\uparrow$ }& \textbf{SSIM $\uparrow$ }& \textbf{LPIPS $\downarrow$ }& \textbf{G-SC $\uparrow$ }&\textbf{G-PQ $\uparrow$ }& \textbf{G-O $\uparrow$ }& \textbf{Latency (s) $\downarrow$ }&\textbf{Speedup $\uparrow$ }\\
        \midrule
         \textbf{Step1X-Edit}~\citep{step1x2025liu}&  -&  -&  -& 8.300 &6.575 &  7.226 &  27.880 &1.000 
\\
         \; \textbf{+ Stepskip}&  24.377 &  0.858 &  0.117 & 8.050 &5.900 &  6.676 &  12.260 &2.274 
\\
         \; \textbf{+ FORA}~\citep{selvaraju2024fora}&  20.406 &  0.763 &  0.224 & 7.175 &6.875 &  6.579 &  14.286 &1.952 
\\
         \; \textbf{+ $\Delta$-DiT}~\citep{chen2024deltadit}&  21.995 &  0.829 &  0.154 & 8.025 &5.975 &  6.695 &  12.685 &2.198 
\\
         \; \textbf{+ TeaCache}~\citep{liu2025teacache}&  25.630 &  0.875 &  0.099 & 8.175 &6.000 &  6.796 &  11.163 &2.498 
\\
         \; \textbf{+ RAS}~\citep{liu2025ras}&  24.302 &  0.844 &  0.141 & 8.275 &5.700 &  6.633 &  15.202 &1.834 
\\
         \; \textbf{+ ToCa}~\citep{zou2024toca}& 22.503 & 0.793 & 0.186 & 7.850 &5.450 & 6.352 & 22.306 &1.250 
\\
        
        \rowcolor[HTML]{EFEFEF}
         \; \textbf{+ Ours (RegionE)}& 27.248 & 0.897 & 0.080 & 8.475 &6.200 & 6.997 & 11.251 &2.478 
\\
         \midrule
         \textbf{Qwen-Image-Edit}~\citep{qwenimage2025wu}& -& -& -& 8.725 &7.150 & 7.629 & 32.156 &1.000 
\\
         \; \textbf{+ Stepskip}
& 26.300 & 0.870 & 0.093 & 8.650 &6.875 & 7.557 & 17.578 &1.829 
\\
         \; \textbf{+ FORA}~\citep{selvaraju2024fora}
& 24.699 & 0.841 & 0.116 & 8.525 &6.675 & 7.389 & 17.839 &1.803 
\\
         \; \textbf{+ $\Delta$-DiT}~\citep{chen2024deltadit}
& 23.827 & 0.799 & 0.133 & 8.425 &6.475 & 7.205 & 17.472 &1.840 
\\
         \; \textbf{+ TeaCache}~\citep{liu2025teacache}
& 26.788 & 0.876 & 0.092 & 8.725 &6.775 & 7.564 & 16.485 &1.951 
\\
         \; \textbf{+ RAS}~\citep{liu2025ras}
& 26.927 & 0.862 & 0.098 & 8.400 &6.625 & 7.192 & 22.357 &1.438 
\\
         \; \textbf{+ ToCa}~\citep{zou2024toca}
& OOM& OOM& OOM& OOM&OOM& OOM& OOM&OOM\\
        
         \rowcolor[HTML]{EFEFEF}
         \; \textbf{+ Ours (RegionE)}& 30.024 & 0.917 & 0.060 & 8.550 &6.900 & 7.415 & 15.671 &2.052 
\\
         \bottomrule[1.5pt]
    \end{tabular}
    }
    \label{tab:11_material_alter}
\end{table}
\begin{table}
    \caption{Comparison of RegionE and other baselines on the subject-add task of GEdit-Bench, evaluated in terms of quality and efficiency.}
    \resizebox{\linewidth}{!}{%
    \begin{tabular}{l|ccc|ccc|cc}
        \toprule[1.5pt]
         \multirow{2}{*}{\textbf{Model}}&  \multicolumn{3}{c|}{\textbf{Against Vanilla}}& \multicolumn{3}{c|}{\textbf{GPT-4o Score}}&  \multicolumn{2}{c}{\textbf{Efficiency}}\\
 & \textbf{PSNR$\uparrow$ }& \textbf{SSIM $\uparrow$ }& \textbf{LPIPS $\downarrow$ }& \textbf{G-SC $\uparrow$ }&\textbf{G-PQ $\uparrow$ }& \textbf{G-O $\uparrow$ }& \textbf{Latency (s) $\downarrow$ }&\textbf{Speedup $\uparrow$ }\\
        \midrule
         \textbf{Step1X-Edit}~\citep{step1x2025liu}&  -&  -&  -& 8.283 &7.950 &  7.905 &  27.912 &1.000 
\\
         \; \textbf{+ Stepskip}&  25.692 &  0.892 &  0.085 & 8.583 &8.083 &  8.142 &  12.290 &2.271 
\\
         \; \textbf{+ FORA}~\citep{selvaraju2024fora}&  21.717 &  0.848 &  0.150 & 6.400 &8.083 &  6.131 &  14.322 &1.949 
\\
         \; \textbf{+ $\Delta$-DiT}~\citep{chen2024deltadit}&  24.203 &  0.868 &  0.099 & 8.017 &8.050 &  7.655 &  12.727 &2.193 
\\
         \; \textbf{+ TeaCache}~\citep{liu2025teacache}&  26.413 &  0.914 &  0.073 & 8.600 &8.067 &  8.177 &  11.204 &2.491 
\\
         \; \textbf{+ RAS}~\citep{liu2025ras}&  25.008 &  0.880 &  0.101 & 8.100 &7.517 &  7.532 &  15.232 &1.832 
\\
         \; \textbf{+ ToCa}~\citep{zou2024toca}& 23.524 & 0.820 & 0.159 & 7.650 &6.950 & 6.939 & 22.062 &1.265 
\\
        
        \rowcolor[HTML]{EFEFEF}
         \; \textbf{+ Ours (RegionE)}& 28.514 & 0.923 & 0.058 & 8.383 &7.950 & 7.858 & 10.528 &2.651 
\\
         \midrule
         \textbf{Qwen-Image-Edit}~\citep{qwenimage2025wu}& -& -& -& 9.117 &8.017 & 8.381 & 32.081 &1.000 
\\
         \; \textbf{+ Stepskip}
& 27.666 & 0.890 & 0.092 & 8.767 &7.950 & 8.146 & 17.532 &1.830 
\\
         \; \textbf{+ FORA}~\citep{selvaraju2024fora}
& 26.871 & 0.879 & 0.093 & 9.017 &7.933 & 8.313 & 17.810 &1.801 
\\
         \; \textbf{+ $\Delta$-DiT}~\citep{chen2024deltadit}
& 25.559 & 0.849 & 0.108 & 8.617 &7.817 & 7.967 & 17.452 &1.838 
\\
         \; \textbf{+ TeaCache}~\citep{liu2025teacache}
& 28.672 & 0.903 & 0.066 & 8.783 &7.933 & 8.099 & 16.422 &1.954 
\\
         \; \textbf{+ RAS}~\citep{liu2025ras}
& 27.398 & 0.891 & 0.081 & 9.100 &7.933 & 8.267 & 22.278 &1.440 
\\
         \; \textbf{+ ToCa}~\citep{zou2024toca}
& OOM& OOM& OOM& OOM&OOM& OOM& OOM&OOM\\
        
         \rowcolor[HTML]{EFEFEF}
         \; \textbf{+ Ours (RegionE)}& 30.763 & 0.938 & 0.050 & 8.983 &8.233 & 8.441 & 15.295 &2.097 
\\
         \bottomrule[1.5pt]
    \end{tabular}
    }
    \label{tab:11_subject_add}
\end{table}
\begin{table}
    \caption{Comparison of RegionE and other baselines on the subject-remove task of GEdit-Bench, evaluated in terms of quality and efficiency.}
    \resizebox{\linewidth}{!}{%
    \begin{tabular}{l|ccc|ccc|cc}
        \toprule[1.5pt]
         \multirow{2}{*}{\textbf{Model}}&  \multicolumn{3}{c|}{\textbf{Against Vanilla}}& \multicolumn{3}{c|}{\textbf{GPT-4o Score}}&  \multicolumn{2}{c}{\textbf{Efficiency}}\\
 & \textbf{PSNR$\uparrow$ }& \textbf{SSIM $\uparrow$ }& \textbf{LPIPS $\downarrow$ }& \textbf{G-SC $\uparrow$ }&\textbf{G-PQ $\uparrow$ }& \textbf{G-O $\uparrow$ }& \textbf{Latency (s) $\downarrow$ }&\textbf{Speedup $\uparrow$ }\\
        \midrule
         \textbf{Step1X-Edit}~\citep{step1x2025liu}&  -&  -&  -& 7.351 &7.947 &  6.973 &  27.954 &1.000 
\\
         \; \textbf{+ Stepskip}&  33.649 &  0.954 &  0.038 & 7.579 &7.684 &  6.969 &  12.300 &2.273 
\\
         \; \textbf{+ FORA}~\citep{selvaraju2024fora}&  30.330 &  0.943 &  0.062 & 5.474 &7.895 &  5.285 &  14.330 &1.951 
\\
         \; \textbf{+ $\Delta$-DiT}~\citep{chen2024deltadit}&  31.847 &  0.948 &  0.047 & 7.930 &7.684 &  7.319 &  12.724 &2.197 
\\
         \; \textbf{+ TeaCache}~\citep{liu2025teacache}&  36.735 &  0.973 &  0.024 & 7.281 &7.737 &  6.841 &  11.213 &2.493 
\\
         \; \textbf{+ RAS}~\citep{liu2025ras}&  32.966 &  0.936 &  0.052 & 7.211 &7.860 &  6.861 &  15.236 &1.835 
\\
         \; \textbf{+ ToCa}~\citep{zou2024toca}& 29.806 & 0.894 & 0.095 & 7.175 &7.088 & 6.481 & 22.378 &1.249 
\\
        
        \rowcolor[HTML]{EFEFEF}
         \; \textbf{+ Ours (RegionE)}& 35.772 & 0.963 & 0.028 & 7.719 &7.737 & 7.182 & 10.453 &2.674 
\\
         \midrule
         \textbf{Qwen-Image-Edit}~\citep{qwenimage2025wu}& -& -& -& 8.965 &8.246 & 8.477 & 32.170 &1.000 
\\
         \; \textbf{+ Stepskip}
& 32.187 & 0.913 & 0.048 & 9.035 &8.298 & 8.558 & 17.572 &1.831 
\\
         \; \textbf{+ FORA}~\citep{selvaraju2024fora}
& 29.288 & 0.865 & 0.072 & 9.175 &7.930 & 8.475 & 17.820 &1.805 
\\
         \; \textbf{+ $\Delta$-DiT}~\citep{chen2024deltadit}
& 27.056 & 0.826 & 0.090 & 8.895 &7.947 & 8.348 & 17.486 &1.840 
\\
         \; \textbf{+ TeaCache}~\citep{liu2025teacache}
& 31.687 & 0.899 & 0.051 & 8.895 &8.228 & 8.441 & 16.434 &1.958 
\\
         \; \textbf{+ RAS}~\citep{liu2025ras}
& 28.440 & 0.876 & 0.080 & 8.842 &8.035 & 8.371 & 22.331 &1.441 
\\
         \; \textbf{+ ToCa}~\citep{zou2024toca}
& OOM& OOM& OOM& OOM&OOM& OOM& OOM&OOM\\
        
         \rowcolor[HTML]{EFEFEF}
         \; \textbf{+ Ours (RegionE)}& 32.122 & 0.925 & 0.052 & 9.333 &8.351 & 8.787 & 15.349 &2.096 
\\
         \bottomrule[1.5pt]
    \end{tabular}
    }
    \label{tab:11_subject_remove}
\end{table}
\begin{table}
    \caption{Comparison of RegionE and other baselines on the style-change task of GEdit-Bench, evaluated in terms of quality and efficiency.}
    \resizebox{\linewidth}{!}{%
    \begin{tabular}{l|ccc|ccc|cc}
        \toprule[1.5pt]
         \multirow{2}{*}{\textbf{Model}}&  \multicolumn{3}{c|}{\textbf{Against Vanilla}}& \multicolumn{3}{c|}{\textbf{GPT-4o Score}}&  \multicolumn{2}{c}{\textbf{Efficiency}}\\
 & \textbf{PSNR$\uparrow$ }& \textbf{SSIM $\uparrow$ }& \textbf{LPIPS $\downarrow$ }& \textbf{G-SC $\uparrow$ }&\textbf{G-PQ $\uparrow$ }& \textbf{G-O $\uparrow$ }& \textbf{Latency (s) $\downarrow$ }&\textbf{Speedup $\uparrow$ }\\
        \midrule
         \textbf{Step1X-Edit}~\citep{step1x2025liu}&  -&  -&  -& 8.150 &6.917 &  7.359 &  27.898 &1.000 
\\
         \; \textbf{+ Stepskip}&  21.064 &  0.828 &  0.185 & 8.183 &6.583 &  7.199 &  12.277 &2.272 
\\
         \; \textbf{+ FORA}~\citep{selvaraju2024fora}&  15.851 &  0.680 &  0.372 & 7.183 &7.017 &  6.883 &  14.300 &1.951 
\\
         \; \textbf{+ $\Delta$-DiT}~\citep{chen2024deltadit}&  18.893 &  0.791 &  0.233 & 8.167 &6.367 &  7.066 &  12.684 &2.200 
\\
         \; \textbf{+ TeaCache}~\citep{liu2025teacache}&  21.695 &  0.857 &  0.156 & 8.000 &6.733 &  7.213 &  11.187 &2.494 
\\
         \; \textbf{+ RAS}~\citep{liu2025ras}&  21.355 &  0.814 &  0.193 & 8.217 &6.400 &  7.108 &  15.217 &1.833 
\\
         \; \textbf{+ ToCa}~\citep{zou2024toca}& 19.819 & 0.760 & 0.250 & 8.283 &6.000 & 6.927 & 22.327 &1.250 
\\
        
        \rowcolor[HTML]{EFEFEF}
         \; \textbf{+ Ours (RegionE)}& 25.449 & 0.900 & 0.102 & 8.267 &6.617 & 7.251 & 11.797 &2.365 
\\
         \midrule
         \textbf{Qwen-Image-Edit}~\citep{qwenimage2025wu}& -& -& -& 8.267 &7.133 & 7.526 & 32.115 &1.000 
\\
         \; \textbf{+ Stepskip}
& 23.954 & 0.805 & 0.139 & 8.067 &7.083 & 7.355 & 17.560 &1.829 
\\
         \; \textbf{+ FORA}~\citep{selvaraju2024fora}
& 21.784 & 0.745 & 0.185 & 8.017 &7.100 & 7.385 & 17.807 &1.804 
\\
         \; \textbf{+ $\Delta$-DiT}~\citep{chen2024deltadit}
& 20.552 & 0.662 & 0.219 & 8.117 &7.033 & 7.395 & 17.455 &1.840 
\\
         \; \textbf{+ TeaCache}~\citep{liu2025teacache}
& 23.137 & 0.816 & 0.152 & 8.300 &7.133 & 7.544 & 16.414 &1.957 
\\
         \; \textbf{+ RAS}~\citep{liu2025ras}
& 24.073 & 0.772 & 0.169 & 7.983 &7.000 & 7.348 & 22.275 &1.442 
\\
         \; \textbf{+ ToCa}~\citep{zou2024toca}
& OOM& OOM& OOM& OOM&OOM& OOM& OOM&OOM\\
        
         \rowcolor[HTML]{EFEFEF}
         \; \textbf{+ Ours (RegionE)}& 27.980 & 0.897 & 0.073 & 8.233 &7.250 & 7.583 & 16.822 &1.909 
\\
         \bottomrule[1.5pt]
    \end{tabular}
    }
    \label{tab:11_style_change}
\end{table}
\begin{table}
    \caption{Comparison of RegionE and other baselines on the tone-transfer task of GEdit-Bench, evaluated in terms of quality and efficiency.}
    \resizebox{\linewidth}{!}{%
    \begin{tabular}{l|ccc|ccc|cc}
        \toprule[1.5pt]
         \multirow{2}{*}{\textbf{Model}}&  \multicolumn{3}{c|}{\textbf{Against Vanilla}}& \multicolumn{3}{c|}{\textbf{GPT-4o Score}}&  \multicolumn{2}{c}{\textbf{Efficiency}}\\
 & \textbf{PSNR$\uparrow$ }& \textbf{SSIM $\uparrow$ }& \textbf{LPIPS $\downarrow$ }& \textbf{G-SC $\uparrow$ }&\textbf{G-PQ $\uparrow$ }& \textbf{G-O $\uparrow$ }& \textbf{Latency (s) $\downarrow$ }&\textbf{Speedup $\uparrow$ }\\
        \midrule
         \textbf{Step1X-Edit}~\citep{step1x2025liu}&  -&  -&  -& 6.950 &7.325 &  6.679 &  27.874 &1.000 
\\
         \; \textbf{+ Stepskip}&  24.744 &  0.899 &  0.122 & 7.200 &7.325 &  6.917 &  12.260 &2.274 
\\
         \; \textbf{+ FORA}~\citep{selvaraju2024fora}&  19.078 &  0.786 &  0.251 & 6.900 &8.125 &  7.088 &  14.293 &1.950 
\\
         \; \textbf{+ $\Delta$-DiT}~\citep{chen2024deltadit}&  22.104 &  0.862 &  0.164 & 7.000 &7.250 &  6.852 &  12.678 &2.199 
\\
         \; \textbf{+ TeaCache}~\citep{liu2025teacache}&  27.915 &  0.933 &  0.072 & 6.825 &7.400 &  6.600 &  11.171 &2.495 
\\
         \; \textbf{+ RAS}~\citep{liu2025ras}&  26.455 &  0.895 &  0.111 & 7.200 &7.000 &  6.688 &  15.187 &1.835 
\\
         \; \textbf{+ ToCa}~\citep{zou2024toca}& 23.954 & 0.840 & 0.159 & 6.500 &6.550 & 5.991 & 22.408 &1.244 
\\
        
        \rowcolor[HTML]{EFEFEF}
         \; \textbf{+ Ours (RegionE)}& 30.860 & 0.945 & 0.064 & 6.900 &7.275 & 6.641 & 11.496 &2.425 
\\
         \midrule
         \textbf{Qwen-Image-Edit}~\citep{qwenimage2025wu}& -& -& -& 8.475 &8.025 & 8.084 & 32.160 &1.000 
\\
         \; \textbf{+ Stepskip}
& 29.715 & 0.862 & 0.092 & 8.150 &8.000 & 7.820 & 17.562 &1.831 
\\
         \; \textbf{+ FORA}~\citep{selvaraju2024fora}
& 27.514 & 0.839 & 0.117 & 8.025 &7.875 & 7.771 & 17.841 &1.803 
\\
         \; \textbf{+ $\Delta$-DiT}~\citep{chen2024deltadit}
& 25.471 & 0.792 & 0.139 & 7.950 &7.725 & 7.592 & 17.462 &1.842 
\\
         \; \textbf{+ TeaCache}~\citep{liu2025teacache}
& 30.064 & 0.910 & 0.061 & 8.375 &8.125 & 8.033 & 16.381 &1.963 
\\
         \; \textbf{+ RAS}~\citep{liu2025ras}
& 29.142 & 0.880 & 0.089 & 8.500 &8.075 & 8.142 & 22.372 &1.437 
\\
         \; \textbf{+ ToCa}~\citep{zou2024toca}
& OOM& OOM& OOM& OOM&OOM& OOM& OOM&OOM\\
        
         \rowcolor[HTML]{EFEFEF}
         \; \textbf{+ Ours (RegionE)}& 34.051 & 0.948 & 0.034 & 8.450 &8.275 & 8.199 & 15.851 &2.029 
\\
         \bottomrule[1.5pt]
    \end{tabular}
    }
    \label{tab:11_tone_transfer}
\end{table}
\begin{table}
    \caption{Comparison of RegionE and other baselines on the subject-replace task of GEdit-Bench, evaluated in terms of quality and efficiency.}
    \resizebox{\linewidth}{!}{%
    \begin{tabular}{l|ccc|ccc|cc}
        \toprule[1.5pt]
         \multirow{2}{*}{\textbf{Model}}&  \multicolumn{3}{c|}{\textbf{Against Vanilla}}& \multicolumn{3}{c|}{\textbf{GPT-4o Score}}&  \multicolumn{2}{c}{\textbf{Efficiency}}\\
 & \textbf{PSNR$\uparrow$ }& \textbf{SSIM $\uparrow$ }& \textbf{LPIPS $\downarrow$ }& \textbf{G-SC $\uparrow$ }&\textbf{G-PQ $\uparrow$ }& \textbf{G-O $\uparrow$ }& \textbf{Latency (s) $\downarrow$ }&\textbf{Speedup $\uparrow$ }\\
        \midrule
         \textbf{Step1X-Edit}~\citep{step1x2025liu}&  -&  -&  -& 8.650 &7.233 &  7.718 &  27.983 &1.000 
\\
         \; \textbf{+ Stepskip}&  25.233 &  0.875 &  0.111 & 8.683 &6.867 &  7.548 &  12.325 &2.270 
\\
         \; \textbf{+ FORA}~\citep{selvaraju2024fora}&  20.594 &  0.831 &  0.189 & 5.833 &6.817 &  5.306 &  14.359 &1.949 
\\
         \; \textbf{+ $\Delta$-DiT}~\citep{chen2024deltadit}&  22.927 &  0.835 &  0.141 & 8.500 &6.733 &  7.345 &  12.766 &2.192 
\\
         \; \textbf{+ TeaCache}~\citep{liu2025teacache}&  25.856 &  0.915 &  0.088 & 8.417 &7.183 &  7.536 &  11.245 &2.488 
\\
         \; \textbf{+ RAS}~\citep{liu2025ras}&  25.072 &  0.888 &  0.116 & 8.250 &6.433 &  6.996 &  15.268 &1.833 
\\
         \; \textbf{+ ToCa}~\citep{zou2024toca}& 23.407 & 0.840 & 0.168 & 8.267 &6.217 & 6.909 & 22.080 &1.267 
\\
        
        \rowcolor[HTML]{EFEFEF}
         \; \textbf{+ Ours (RegionE)}& 28.654 & 0.935 & 0.064 & 8.517 &7.167 & 7.585 & 10.647 &2.628 
\\
         \midrule
         \textbf{Qwen-Image-Edit}~\citep{qwenimage2025wu}& inf& 1.000 & 0.000 & 8.883 &7.683 & 8.136 & 32.161 &1.000 
\\
         \; \textbf{+ Stepskip}
& 26.344 & 0.897 & 0.076 & 8.783 &7.733 & 8.128 & 17.575 &1.830 
\\
         \; \textbf{+ FORA}~\citep{selvaraju2024fora}
& 24.578 & 0.864 & 0.104 & 8.600 &7.550 & 7.930 & 17.836 &1.803 
\\
         \; \textbf{+ $\Delta$-DiT}~\citep{chen2024deltadit}
& 23.745 & 0.829 & 0.120 & 8.450 &7.267 & 7.687 & 17.496 &1.838 
\\
         \; \textbf{+ TeaCache}~\citep{liu2025teacache}
& 25.993 & 0.891 & 0.084 & 8.700 &7.733 & 8.120 & 16.391 &1.962 
\\
         \; \textbf{+ RAS}~\citep{liu2025ras}
& 25.579 & 0.881 & 0.095 & 8.867 &7.400 & 7.996 & 22.341 &1.440 
\\
         \; \textbf{+ ToCa}~\citep{zou2024toca}
& OOM& OOM& OOM& OOM&OOM& OOM& OOM&OOM\\
        
         \rowcolor[HTML]{EFEFEF}
         \; \textbf{+ Ours (RegionE)}& 29.388 & 0.938 & 0.047 & 8.967 &7.767 & 8.242 & 15.446 &2.082 
\\
         \bottomrule[1.5pt]
    \end{tabular}
    }
    \label{tab:11_subject_replace}
\end{table}
\begin{table}
    \caption{Comparison of RegionE and other baselines on the text-change task of GEdit-Bench, evaluated in terms of quality and efficiency.}
    \resizebox{\linewidth}{!}{%
    \begin{tabular}{l|ccc|ccc|cc}
        \toprule[1.5pt]
         \multirow{2}{*}{\textbf{Model}}&  \multicolumn{3}{c|}{\textbf{Against Vanilla}}& \multicolumn{3}{c|}{\textbf{GPT-4o Score}}&  \multicolumn{2}{c}{\textbf{Efficiency}}\\
 & \textbf{PSNR$\uparrow$ }& \textbf{SSIM $\uparrow$ }& \textbf{LPIPS $\downarrow$ }& \textbf{G-SC $\uparrow$ }&\textbf{G-PQ $\uparrow$ }& \textbf{G-O $\uparrow$ }& \textbf{Latency (s) $\downarrow$ }&\textbf{Speedup $\uparrow$ }\\
        \midrule
         \textbf{Step1X-Edit}~\citep{step1x2025liu}&  -&  -&  -& 8.293 &8.091 &  7.900 &  28.027 &1.000 
\\
         \; \textbf{+ Stepskip}&  30.069 &  0.955 &  0.032 & 8.222 &8.192 &  7.951 &  12.331 &2.273 
\\
         \; \textbf{+ FORA}~\citep{selvaraju2024fora}&  26.368 &  0.941 &  0.049 & 7.000 &7.899 &  6.926 &  14.375 &1.950 
\\
         \; \textbf{+ $\Delta$-DiT}~\citep{chen2024deltadit}&  28.615 &  0.953 &  0.032 & 8.515 &8.222 &  8.171 &  12.768 &2.195 
\\
         \; \textbf{+ TeaCache}~\citep{liu2025teacache}&  31.420 &  0.967 &  0.023 & 8.222 &8.192 &  7.925 &  11.254 &2.491 
\\
         \; \textbf{+ RAS}~\citep{liu2025ras}&  28.434 &  0.939 &  0.042 & 7.929 &7.970 &  7.649 &  15.270 &1.835 
\\
         \; \textbf{+ ToCa}~\citep{zou2024toca}& 26.305 & 0.902 & 0.078 & 7.949 &7.707 & 7.609 & 21.723 &1.290 
\\
        
        \rowcolor[HTML]{EFEFEF}
         \; \textbf{+ Ours (RegionE)}& 32.404 & 0.968 & 0.020 & 8.212 &8.242 & 8.002 & 10.237 &2.738 
\\
         \midrule
         \textbf{Qwen-Image-Edit}~\citep{qwenimage2025wu}& -& -& -& 9.192 &8.394 & 8.606 & 32.071 &1.000 
\\
         \; \textbf{+ Stepskip}
& 29.577 & 0.929 & 0.047 & 8.828 &8.222 & 8.202 & 17.519 &1.831 
\\
         \; \textbf{+ FORA}~\citep{selvaraju2024fora}
& 27.408 & 0.909 & 0.061 & 8.818 &8.303 & 8.192 & 17.790 &1.803 
\\
         \; \textbf{+ $\Delta$-DiT}~\citep{chen2024deltadit}
& 25.837 & 0.881 & 0.072 & 8.879 &8.333 & 8.259 & 17.432 &1.840 
\\
         \; \textbf{+ TeaCache}~\citep{liu2025teacache}
& 29.126 & 0.932 & 0.047 & 8.778 &8.222 & 8.184 & 16.539 &1.939 
\\
         \; \textbf{+ RAS}~\citep{liu2025ras}
& 26.732 & 0.912 & 0.061 & 8.889 &8.010 & 8.187 & 22.302 &1.438 
\\
         \; \textbf{+ ToCa}~\citep{zou2024toca}
& OOM& OOM& OOM& OOM&OOM& OOM& OOM&OOM\\
        
         \rowcolor[HTML]{EFEFEF}
         \; \textbf{+ Ours (RegionE)}& 31.357 & 0.950 & 0.033 & 8.838 &8.313 & 8.260 & 14.813 &2.165 
\\
         \bottomrule[1.5pt]
    \end{tabular}
    }
    \label{tab:11_text_change}
\end{table}
\begin{table}
    \caption{Comparison of RegionE and other baselines on the background-change task of GEdit-Bench, evaluated in terms of quality and efficiency.}
    \resizebox{\linewidth}{!}{%
    \begin{tabular}{l|ccc|ccc|cc}
        \toprule[1.5pt]
         \multirow{2}{*}{\textbf{Model}}&  \multicolumn{3}{c|}{\textbf{Against Vanilla}}& \multicolumn{3}{c|}{\textbf{GPT-4o Score}}&  \multicolumn{2}{c}{\textbf{Efficiency}}\\
 & \textbf{PSNR$\uparrow$ }& \textbf{SSIM $\uparrow$ }& \textbf{LPIPS $\downarrow$ }& \textbf{G-SC $\uparrow$ }&\textbf{G-PQ $\uparrow$ }& \textbf{G-O $\uparrow$ }& \textbf{Latency (s) $\downarrow$ }&\textbf{Speedup $\uparrow$ }\\
        \midrule
         \textbf{Step1X-Edit}~\citep{step1x2025liu}&  -&  -&  -& 8.575 &7.175 &  7.722 &  27.886 &1.000 
\\
         \; \textbf{+ Stepskip}&  21.011 &  0.812 &  0.218 & 8.625 &6.975 &  7.635 &  12.267 &2.273 
\\
         \; \textbf{+ FORA}~\citep{selvaraju2024fora}&  15.897 &  0.719 &  0.372 & 6.500 &7.125 &  6.104 &  14.285 &1.952 
\\
         \; \textbf{+ $\Delta$-DiT}~\citep{chen2024deltadit}&  18.641 &  0.781 &  0.271 & 8.375 &6.625 &  7.338 &  12.687 &2.198 
\\
         \; \textbf{+ TeaCache}~\citep{liu2025teacache}&  23.549 &  0.866 &  0.151 & 8.375 &6.725 &  7.379 &  11.163 &2.498 
\\
         \; \textbf{+ RAS}~\citep{liu2025ras}&  25.468 &  0.840 &  0.169 & 8.425 &6.725 &  7.372 &  15.206 &1.834 
\\
         \; \textbf{+ ToCa}~\citep{zou2024toca}& 22.925 & 0.768 & 0.255 & 8.200 &6.175 & 6.995 & 22.631 &1.232 
\\
        
        \rowcolor[HTML]{EFEFEF}
         \; \textbf{+ Ours (RegionE)}& 29.076 & 0.917 & 0.091 & 8.500 &7.125 & 7.675 & 11.324 &2.463 
\\
         \midrule
         \textbf{Qwen-Image-Edit}~\citep{qwenimage2025wu}& -& -& -& 9.125 &8.200 & 8.603 & 32.221 &1.000 
\\
         \; \textbf{+ Stepskip}
& 25.088 & 0.854 & 0.133 & 9.175 &7.975 & 8.511 & 17.603 &1.830 
\\
         \; \textbf{+ FORA}~\citep{selvaraju2024fora}
& 22.473 & 0.802 & 0.184 & 8.775 &7.875 & 8.252 & 17.815 &1.809 
\\
         \; \textbf{+ $\Delta$-DiT}~\citep{chen2024deltadit}
& 21.263 & 0.750 & 0.210 & 8.825 &7.850 & 8.281 & 17.552 &1.836 
\\
         \; \textbf{+ TeaCache}~\citep{liu2025teacache}
& 24.016 & 0.852 & 0.147 & 8.850 &7.950 & 8.281 & 16.492 &1.954 
\\
         \; \textbf{+ RAS}~\citep{liu2025ras}
& 26.550 & 0.858 & 0.128 & 9.100 &7.450 & 8.153 & 22.411 &1.438 
\\
         \; \textbf{+ ToCa}~\citep{zou2024toca}
& OOM& OOM& OOM& OOM&OOM& OOM& OOM&OOM\\
        
         \rowcolor[HTML]{EFEFEF}
         \; \textbf{+ Ours (RegionE)}& 30.462 & 0.939 & 0.053 & 9.175 &8.050 & 8.547 & 16.694 &1.930 
\\
         \bottomrule[1.5pt]
    \end{tabular}
    }
    \label{tab:11_background_change}
\end{table}

\begin{table}
    \caption{Comparison of RegionE and other baselines on the Character Reference task of KontextBench, evaluated in terms of quality and efficiency.}
    \resizebox{\linewidth}{!}{%
    \begin{tabular}{l|ccc|ccc|cc}
        \toprule[1.5pt]
         \multirow{2}{*}{\textbf{Model}}&  \multicolumn{3}{c|}{\textbf{Against Vanilla}}& \multicolumn{3}{c|}{\textbf{GPT-4o Score}}&  \multicolumn{2}{c}{\textbf{Efficiency}}\\
 & \textbf{PSNR$\uparrow$ }& \textbf{SSIM $\uparrow$ }& \textbf{LPIPS $\downarrow$ }& \textbf{G-SC $\uparrow$ }&\textbf{G-PQ $\uparrow$ }& \textbf{G-O $\uparrow$ }& \textbf{Latency (s) $\downarrow$ }&\textbf{Speedup $\uparrow$ }\\
        \midrule
         \textbf{FLUX.1 Kontext}~\citep{fluxkontext2025labs}& -& -& -& 7.549 &6.642 & 6.664 & 14.677 &1.000 
\\
         \; \textbf{+ Stepskip}
& 18.793 & 0.730 & 0.238 & 7.741 &6.803 & 6.917 & 8.502 &1.726 
\\
         \; \textbf{+ FORA}~\citep{selvaraju2024fora}
& 17.898 & 0.697 & 0.275 & 7.617 &6.788 & 6.813 & 7.494 &1.958 
\\
         \; \textbf{+ $\Delta$-DiT}~\citep{chen2024deltadit}
& 15.560 & 0.604 & 0.387 & 7.668 &6.451 & 6.704 & 6.737 &2.178 
\\
         \; \textbf{+ TeaCache}~\citep{liu2025teacache}
& 20.313 & 0.770 & 0.197 & 7.865 &6.565 & 6.842 & 6.271 &2.341 
\\
         \; \textbf{+ RAS}~\citep{liu2025ras}
& 21.320 & 0.752 & 0.214 & 7.632 &6.352 & 6.657 & 8.211 &1.788 
\\
         \; \textbf{+ ToCa}~\citep{zou2024toca}
& 19.596 & 0.679 & 0.298 & 7.570 &6.047 & 6.454 & 11.279 &1.301 
\\
        
        \rowcolor[HTML]{EFEFEF}
         \; \textbf{+ Ours (RegionE)}& 26.980 & 0.880 & 0.086 & 7.637 &6.611 & 6.715 & 6.406 &2.291 
\\
         \bottomrule[1.5pt]
    \end{tabular}
    }
    \label{tab:5_CR}
\end{table}
\begin{table}
    \caption{Comparison of RegionE and other baselines on the Instruction Editing-Global task of KontextBench, evaluated in terms of quality and efficiency.}
    \resizebox{\linewidth}{!}{%
    \begin{tabular}{l|ccc|ccc|cc}
        \toprule[1.5pt]
         \multirow{2}{*}{\textbf{Model}}&  \multicolumn{3}{c|}{\textbf{Against Vanilla}}& \multicolumn{3}{c|}{\textbf{GPT-4o Score}}&  \multicolumn{2}{c}{\textbf{Efficiency}}\\
 & \textbf{PSNR$\uparrow$ }& \textbf{SSIM $\uparrow$ }& \textbf{LPIPS $\downarrow$ }& \textbf{G-SC $\uparrow$ }&\textbf{G-PQ $\uparrow$ }& \textbf{G-O $\uparrow$ }& \textbf{Latency (s) $\downarrow$ }&\textbf{Speedup $\uparrow$ }\\
        \midrule
         \textbf{FLUX.1 Kontext}~\citep{fluxkontext2025labs}& -& -& -& 7.023 &6.798 & 6.380 & 14.688 &1.000 
\\
         \; \textbf{+ Stepskip}
& 23.957 & 0.797 & 0.132 & 7.000 &6.870 & 6.435 & 8.516 &1.725 
\\
         \; \textbf{+ FORA}~\citep{selvaraju2024fora}
& 22.611 & 0.760 & 0.159 & 7.092 &6.840 & 6.497 & 7.506 &1.957 
\\
         \; \textbf{+ $\Delta$-DiT}~\citep{chen2024deltadit}
& 18.687 & 0.659 & 0.252 & 7.073 &6.882 & 6.574 & 6.754 &2.175 
\\
         \; \textbf{+ TeaCache}~\citep{liu2025teacache}
& 27.206 & 0.842 & 0.101 & 7.294 &6.885 & 6.626 & 6.251 &2.350 
\\
         \; \textbf{+ RAS}~\citep{liu2025ras}
& 24.845 & 0.778 & 0.157 & 7.302 &6.866 & 6.668 & 8.221 &1.787 
\\
         \; \textbf{+ ToCa}~\citep{zou2024toca}
& 23.030 & 0.711 & 0.218 & 7.179 &6.588 & 6.483 & 11.412 &1.287 
\\
        
        \rowcolor[HTML]{EFEFEF}
         \; \textbf{+ Ours (RegionE)}& 30.403 & 0.886 & 0.071 & 7.126 &6.943 & 6.572 & 6.379 &2.303 
\\
         \bottomrule[1.5pt]
    \end{tabular}
    }
    \label{tab:5_IEG}
\end{table}
\begin{table}
    \caption{Comparison of RegionE and other baselines on the Instruction Editing-Local task of KontextBench, evaluated in terms of quality and efficiency.}
    \resizebox{\linewidth}{!}{%
    \begin{tabular}{l|ccc|ccc|cc}
        \toprule[1.5pt]
         \multirow{2}{*}{\textbf{Model}}&  \multicolumn{3}{c|}{\textbf{Against Vanilla}}& \multicolumn{3}{c|}{\textbf{GPT-4o Score}}&  \multicolumn{2}{c}{\textbf{Efficiency}}\\
 & \textbf{PSNR$\uparrow$ }& \textbf{SSIM $\uparrow$ }& \textbf{LPIPS $\downarrow$ }& \textbf{G-SC $\uparrow$ }&\textbf{G-PQ $\uparrow$ }& \textbf{G-O $\uparrow$ }& \textbf{Latency (s) $\downarrow$ }&\textbf{Speedup $\uparrow$ }\\
        \midrule
         \textbf{FLUX.1 Kontext}~\citep{fluxkontext2025labs}& -& -& -& 6.779 &6.909 & 5.817 & 14.677 &1.000 
\\
         \; \textbf{+ Stepskip}
& 31.147 & 0.913 & 0.058 & 6.839 &6.887 & 5.872 & 8.510 &1.725 
\\
         \; \textbf{+ FORA}~\citep{selvaraju2024fora}
& 29.279 & 0.895 & 0.072 & 6.800 &6.901 & 5.873 & 7.491 &1.959 
\\
         \; \textbf{+ $\Delta$-DiT}~\citep{chen2024deltadit}
& 23.390 & 0.824 & 0.130 & 6.822 &6.829 & 5.846 & 6.751 &2.174 
\\
         \; \textbf{+ TeaCache}~\citep{liu2025teacache}
& 33.341 & 0.938 & 0.040 & 6.942 &6.800 & 5.896 & 6.113 &2.401 
\\
         \; \textbf{+ RAS}~\citep{liu2025ras}
& 30.088 & 0.907 & 0.063 & 6.945 &6.921 & 5.972 & 8.219 &1.786 
\\
         \; \textbf{+ ToCa}~\citep{zou2024toca}
& 26.996 & 0.855 & 0.112 & 6.851 &6.635 & 5.790 & 11.231 &1.307 
\\
        
        \rowcolor[HTML]{EFEFEF}
         \; \textbf{+ Ours (RegionE)}& 36.334 & 0.959 & 0.025 & 6.889 &6.880 & 5.917 & 5.799 &2.531 
\\
         \bottomrule[1.5pt]
    \end{tabular}
    }
    \label{tab:5_IEL}
\end{table}
\begin{table}
    \caption{Comparison of RegionE and other baselines on the Style Reference task of KontextBench, evaluated in terms of quality and efficiency.}
    \resizebox{\linewidth}{!}{%
    \begin{tabular}{l|ccc|ccc|cc}
        \toprule[1.5pt]
         \multirow{2}{*}{\textbf{Model}}&  \multicolumn{3}{c|}{\textbf{Against Vanilla}}& \multicolumn{3}{c|}{\textbf{GPT-4o Score}}&  \multicolumn{2}{c}{\textbf{Efficiency}}\\
 & \textbf{PSNR$\uparrow$ }& \textbf{SSIM $\uparrow$ }& \textbf{LPIPS $\downarrow$ }& \textbf{G-SC $\uparrow$ }&\textbf{G-PQ $\uparrow$ }& \textbf{G-O $\uparrow$ }& \textbf{Latency (s) $\downarrow$ }&\textbf{Speedup $\uparrow$ }\\
        \midrule
         \textbf{FLUX.1 Kontext}~\citep{fluxkontext2025labs}& -& -& -& 6.810 &6.556 & 6.331 & 14.684 &1.000 
\\
         \; \textbf{+ Stepskip}
& 18.606 & 0.678 & 0.290 & 6.333 &6.381 & 5.947 & 8.501 &1.727 
\\
         \; \textbf{+ FORA}~\citep{selvaraju2024fora}
& 17.508 & 0.631 & 0.329 & 6.222 &6.413 & 5.667 & 7.476 &1.964 
\\
         \; \textbf{+ $\Delta$-DiT}~\citep{chen2024deltadit}
& 14.639 & 0.525 & 0.450 & 6.397 &6.873 & 6.108 & 6.731 &2.182 
\\
         \; \textbf{+ TeaCache}~\citep{liu2025teacache}
& 19.781 & 0.712 & 0.264 & 6.444 &6.286 & 5.832 & 6.261 &2.345 
\\
         \; \textbf{+ RAS}~\citep{liu2025ras}
& 19.481 & 0.638 & 0.343 & 6.603 &6.381 & 6.031 & 8.202 &1.790 
\\
         \; \textbf{+ ToCa}~\citep{zou2024toca}
& 18.245 & 0.553 & 0.439 & 6.000 &6.175 & 5.668 & 11.480 &1.279 
\\
        
        \rowcolor[HTML]{EFEFEF}
         \; \textbf{+ Ours (RegionE)}& 24.433 & 0.811 & 0.165 & 6.921 &6.571 & 6.277 & 6.411 &2.291 
\\
         \bottomrule[1.5pt]
    \end{tabular}
    }
    \label{tab:5_SR}
\end{table}
\begin{table}
    \caption{Comparison of RegionE and other baselines on the Text Editing task of KontextBench, evaluated in terms of quality and efficiency.}
    \resizebox{\linewidth}{!}{%
    \begin{tabular}{l|ccc|ccc|cc}
        \toprule[1.5pt]
         \multirow{2}{*}{\textbf{Model}}&  \multicolumn{3}{c|}{\textbf{Against Vanilla}}& \multicolumn{3}{c|}{\textbf{GPT-4o Score}}&  \multicolumn{2}{c}{\textbf{Efficiency}}\\
 & \textbf{PSNR$\uparrow$ }& \textbf{SSIM $\uparrow$ }& \textbf{LPIPS $\downarrow$ }& \textbf{G-SC $\uparrow$ }&\textbf{G-PQ $\uparrow$ }& \textbf{G-O $\uparrow$ }& \textbf{Latency (s) $\downarrow$ }&\textbf{Speedup $\uparrow$ }\\
        \midrule
         \textbf{FLUX.1 Kontext}~\citep{fluxkontext2025labs}& -& -& -& 7.826 &7.913 & 7.295 & 14.697 &1.000 
\\
         \; \textbf{+ Stepskip}
& 30.950 & 0.943 & 0.033 & 7.717 &7.750 & 7.142 & 8.535 &1.722 
\\
         \; \textbf{+ FORA}~\citep{selvaraju2024fora}
& 28.976 & 0.915 & 0.044 & 7.696 &7.543 & 7.067 & 7.520 &1.954 
\\
         \; \textbf{+ $\Delta$-DiT}~\citep{chen2024deltadit}
& 23.931 & 0.839 & 0.085 & 7.315 &7.554 & 6.823 & 6.779 &2.168 
\\
         \; \textbf{+ TeaCache}~\citep{liu2025teacache}
& 31.283 & 0.955 & 0.026 & 7.620 &7.696 & 7.076 & 6.290 &2.336 
\\
         \; \textbf{+ RAS}~\citep{liu2025ras}
& 27.504 & 0.913 & 0.048 & 7.598 &7.402 & 6.971 & 8.238 &1.784 
\\
         \; \textbf{+ ToCa}~\citep{zou2024toca}
& 25.342 & 0.857 & 0.093 & 7.326 &7.500 & 6.791 & 11.206 &1.312 
\\
        
        \rowcolor[HTML]{EFEFEF}
         \; \textbf{+ Ours (RegionE)}& 34.141 & 0.962 & 0.018 & 7.815 &7.761 & 7.211 & 5.767 &2.548 
\\
         \bottomrule[1.5pt]
    \end{tabular}
    }
    \label{tab:5_TE}
\end{table}


\end{document}